%% file: arab_legal_eval.tex
\pdfoutput=1
\documentclass[11pt]{article}

\usepackage{arabtex}
\usepackage{utf8}
\usepackage{wrapfig}
\setcode{utf8}
\usepackage{multirow}
\usepackage{graphicx}
\usepackage{tabularx}
\usepackage{makecell}

\usepackage{enumitem}
\usepackage[preprint]{acl}
\usepackage{amsmath}
\usepackage{hyperref}
\usepackage{times}
\usepackage{latexsym}
\usepackage{listings}
\usepackage{caption}
\usepackage[T1]{fontenc}
\usepackage[utf8]{inputenc}

\usepackage{microtype}
\usepackage{multicol}

\usepackage[arabic, english]{babel}

\usepackage{inconsolata}
\usepackage{booktabs}
\usepackage{graphicx}

\usepackage{float}    % Required for precise placement of figures
\usepackage{multirow}
\usepackage{url}

\title{ArabLegalEval: A Multitask Benchmark for Assessing Arabic Legal Knowledge in Large Language Models}

\author{
 \textbf{Faris Hijazi\textsuperscript{1}},
 \textbf{Somayah AlHarbi\textsuperscript{1}},
 \textbf{Abdulaziz AlHussein\textsuperscript{1}},
 \textbf{Harethah Abu Shairah\textsuperscript{2}},
\\
 \textbf{Reem AlZahrani\textsuperscript{2}},
 \textbf{Hebah AlShamlan\textsuperscript{1}},
 \textbf{Omar Knio\textsuperscript{2}},
 \textbf{George Turkiyyah\textsuperscript{2}}
\\
 \textsuperscript{1}THIQAH,
 \textsuperscript{2}KAUST
}

\begin{document}

\maketitle
\begin{abstract}
The rapid advancements in Large Language Models (LLMs) have led to significant improvements in various natural language processing tasks. However, the evaluation of LLMs' legal knowledge, particularly in non-English languages such as Arabic, remains under-explored. To address this gap, we introduce \verb|ArabLegalEval|, a multitask benchmark dataset for assessing the Arabic legal knowledge of LLMs.
Inspired by the \verb|MMLU| and \verb|LegalBench| datasets, \verb|ArabLegalEval| consists of multiple tasks sourced from Saudi legal documents and synthesized questions. In this work, we aim to analyze the capabilities required to solve legal problems in Arabic and benchmark the performance of state-of-the-art LLMs. We explore the impact of in-context learning and investigate various evaluation methods. Additionally, we explore workflows for generating questions with automatic validation to enhance the dataset's quality. We benchmark multilingual and Arabic-centric LLMs, such as \verb|GPT-4| and \verb|Jais|, respectively. We also share our methodology for creating the dataset and validation, which can be generalized to other domains.
We hope to accelerate AI research in the Arabic Legal domain by releasing the ArabLegalEval dataset and code: \href{https://github.com/Thiqah/ArabLegalEval}{https://github.com/Thiqah/ArabLegalEval}
% We observe that depending on the format of the question, LLMs achieve drastically different scores for questions requiring the same knowledge. even State-of-the-art LLMs like GPT4 are lacking in open-ended question answering in new domains even when the required context is provided, and it's not far ahead of open source models. 
\end{abstract}

% \newpage 
% Section 1: Introduction
\input{latex/intro}

% \newpage
% Section 2: Related Work
\input{latex/relatedwork}

% \newpage
% Section 3: Data Sources

\input{latex/arabLegalEval}   % section retitled in file
% 

% \newpage
% Section 4: Benckmark Tasks
\section{Benckmark Tasks}
In this section, we describe the three broad task categories in the benchmark, including 10,000+ MCQs from the native Arabic MoJ and BoE documents, a set of QA from these documents, and a quality translation of a subset of the English \verb|LegalBench| benchmark related to consumer contracts and privacy policies. See Figure \ref{fig:data sources}. We believe that the mixture of questions from native Arabic documents and translated questions gives us a somewhat diverse set of tasks and allows the benchmark to test a broader set of capabilities. In addition, this allows us to test the observation that, with increasing model scale, multilingual LLMs can display reasoning abilities and semantic judgment in Arabic as well as in they do in English \cite{shi23}.
The quantitative details and human performance baseline are presented in Appendix \ref{sec:quant-analysis}.

\input{latex/MCQ}

\input{latex/QA}

% \input{latex/translation} 

\input{latex/arlegalbench}    % this is an edit of the translation file

% \newpage 
% Section 5: Evaluation
\section{Evaluation and results}
\input{latex/mcq_eval}

\input{latex/qa_eval}

\input{latex/eval_somayah.tex}

% \subsection{Results}
% How good is the current crop of LLMs at solving the tasks of this benchmark?  
% GPT4, CLAUDE3, Aya, Jais, GPT3.5
\section{Limitations}

The \verb|ArabLegalEval| benchmark currently relies heavily on Saudi Arabian legal documents, with some tasks translated from universal benchmarks. Including documents from more Arabic-speaking countries would improve geographic representation. Our study did not evaluate all models, which limits generalizability; future work should include a broader range of models. Limited access to legal experts affected validation depth; involving more experts would improve quality control. The dataset lacks granular categorization, such as task-specific prior knowledge, document origin, and AI-generated content labels. Adding more granular metadata and task categories would aid nuanced model training and evaluation.

% \newpage
% Section 6: Conclusions
% \input{latex/conclusion}
\input{latex/conclusions2}

% \subsection{Future work and limitations}
% - we should run more LLMs
% - TODO:

%\begin{table*}[h]
%\centering
%\begin{tabular}{|l |l |l|} \hline 
%Model & Language & \# parameters \\ \hline 
%GPT 4o & Multilingual & Undisclosed \\ \hline 
%GPT 4 (0125-preview) & Multilingual & Undisclosed \\ \hline 
%GPT 3.5 turbo 16k (0613) & Multilingual & Undisclosed \\ \hline 
%meta-llama/Meta-Llama-3-8B-Instruct & Multilingual & 8B \\ \hline 
%meta-llama/Meta-Llama-3-70B-Instruct & Multilingual & 70B \\ \hline 
%core42/jais-30b-chat-v3 & Arabic & 30B \\ \hline 
%core42/jais-13b-chat & Arabic & 13B \\ \hline 
%CohereForAI/c4ai-command-r-v01 & Multilingual & 35B \\ \hline 
%CohereForAI/c4ai-command-r-plus & Multilingual & 104B \\ \hline 
%CohereForAI/aya-101 & Multilingual & 13B \\ \hline 
%Claude 3 opus (20240229) & Multilingual & 137B \\ \hline

%\end{tabular}
%\end{table*}

\section*{Acknowledgments}
We would like to express our sincere gratitude to Yusra Alonaizan for her contributions in envisioning this project, and dedicated team of six members—Lama Alsaif, Alanoud Alangari, Alaa Bazaid, and Faisal Alessa for gathering and curating the data sources. 
% Her visionary idea of developing a large pre-trained model for legal tasks predated the release of \verb|ChatGPT| (\verb|GPT-3.5| and \verb|GPT-4|), significantly shaping the direction of our research. 
Additionally, we extend our thanks to the legal team at THIQAH for their valuable time in reviewing the labeled data, we are grateful to Norah AlHussein, an external legal expert, for her meticulous double-review of the labeled data and the translated LegalBench dataset, as well as for her clarifications and support in explaining the legal data.

\nocite{gao23_eval-harness, alghamdi23, jin23_inenglish, mukherjee21_fewshot}
\nocite{abdelali24_larabench, dalvi23_llmbench}
\nocite{alghamdi22_armath}
\nocite{kolt2022predicting,Zheng-et-al-judging,ft-judge-models,prometheus}

\bibliography{arab_legal_eval}

\clearpage

\appendix

\input{latex/appendix_model_info}
\newpage
\input{latex/appendix_exfigures}
\newpage
\input{latex/appendix_MCQs}
\newpage
\input{latex/appendix_rouge}

\newpage
\input{latex/appendix_qa}

% \newpage

\begin{figure}[h!]
    \centering
    \includegraphics[width=0.5\textwidth]{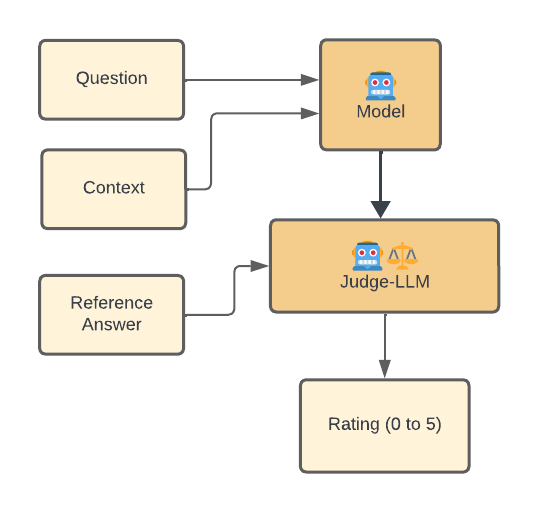}
    \caption{LLM as a judge}
    \label{fig:latex/llm-judge.png}
\end{figure}
\newpage
\section{Data statistics}

\begin{figure}[H]
    \centering
    \includegraphics[width=1\columnwidth]{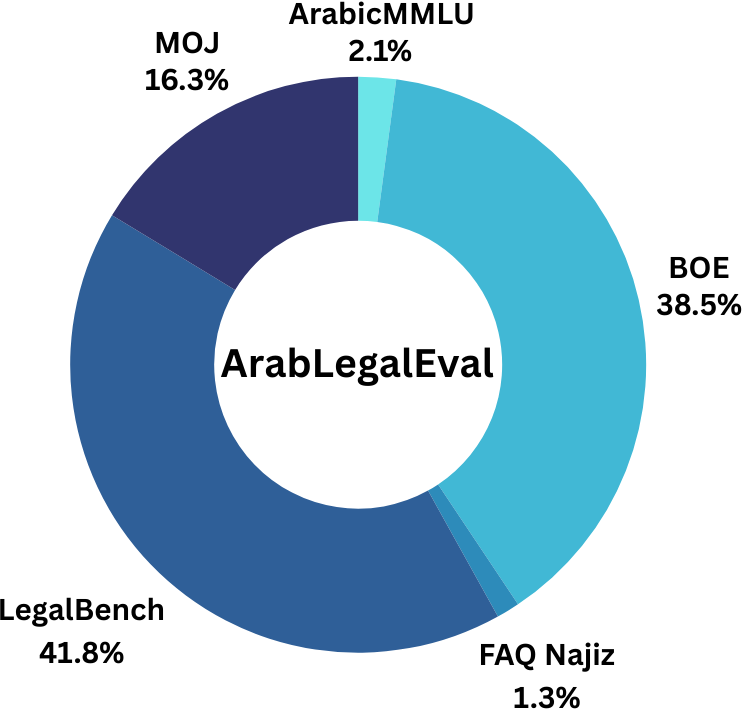}
    \caption{Source documents used to create ArabLegalEval and their percentages}
    \label{fig:sunburstchart}
\end{figure}

\begin{table}[H]
\centering
\begin{tabular}{p{0.25\linewidth} p{0.5\linewidth} p{0.25\linewidth}}
\toprule
\textbf{Source} & \textbf{Category} & \textbf{Value} \\
\midrule
\multirow{4}{2.5cm}{LegalBench} & Privacy Policy Entailment & 4385 \\
 & Privacy Policy QA & 10931 \\
 & Contracts QA & 88 \\
 & Consumer Contracts QA & 400 \\
\midrule
\multirow{2}{2.5cm}{BOE} & Rules Count & 448 \\
 & Rules Count - Subject & 14134 \\
\midrule
\multirow{3}{2.5cm}{MOJ} & Regulations & 67 \\
 & Regulations - Subjects & 5720 \\
 & Circular & 388 \\
\midrule
FAQ Najiz & FAQ Najiz & 492 \\
\midrule
ArabicMMLU & ArabicMMLU & 800 \\
\bottomrule
\end{tabular}
\caption{Tasks counts in ArabLegalEval and their source documents.}
\end{table}
\input{latex/appendix_arbmmlu}

\input{latex/appendix_ArabicLB_Results}
\input{latex/appendix_LegalBench_Prompts}
\input{latex/appendix_quantAnalysis}

\end{document}

%% file: latex/intro.tex
\section{Introduction}

The development of LLMs has revolutionized various fields by enhancing natural language understanding and generation capabilities. However, the applicability and performance of these models in specialized domains, such as legal contexts, specially in low- and medium-resource languages such as Arabic, remain active research areas. In this paper, we report on ongoing work for evaluating the proficiency of large language models in understanding and processing Arabic legal texts. Given the complexity and richness of legal language, especially in Arabic, it is crucial to develop benchmarks that accurately assess the models' capabilities in this domain in order to guide model development. 
One of the key motivations for this work and benchmark is to find out the current state of LLMs in the Arabic Legal domain. Thus, we benchmark a wide range of LLMs from proprietary multilingual LLMs, such as \verb|GPT-4| \cite{openai2024gpt4}, to open-source Arabic-centric LLMs, such as \verb|Jais| \cite{jais23}.

% The rapid advancements in large language models (LLMs) have significantly improved natural language processing tasks, yet the evaluation of these models' legal knowledge, particularly in non-English languages such as Arabic, remains underexplored. Initially, we considered developing an Arabic Legal LLM to handle complex legal tasks. However, the fast-paced evolution of AI means that any new model could quickly become obsolete.

\begin{figure}[t]
    \centering
    \includegraphics[width=1\columnwidth]{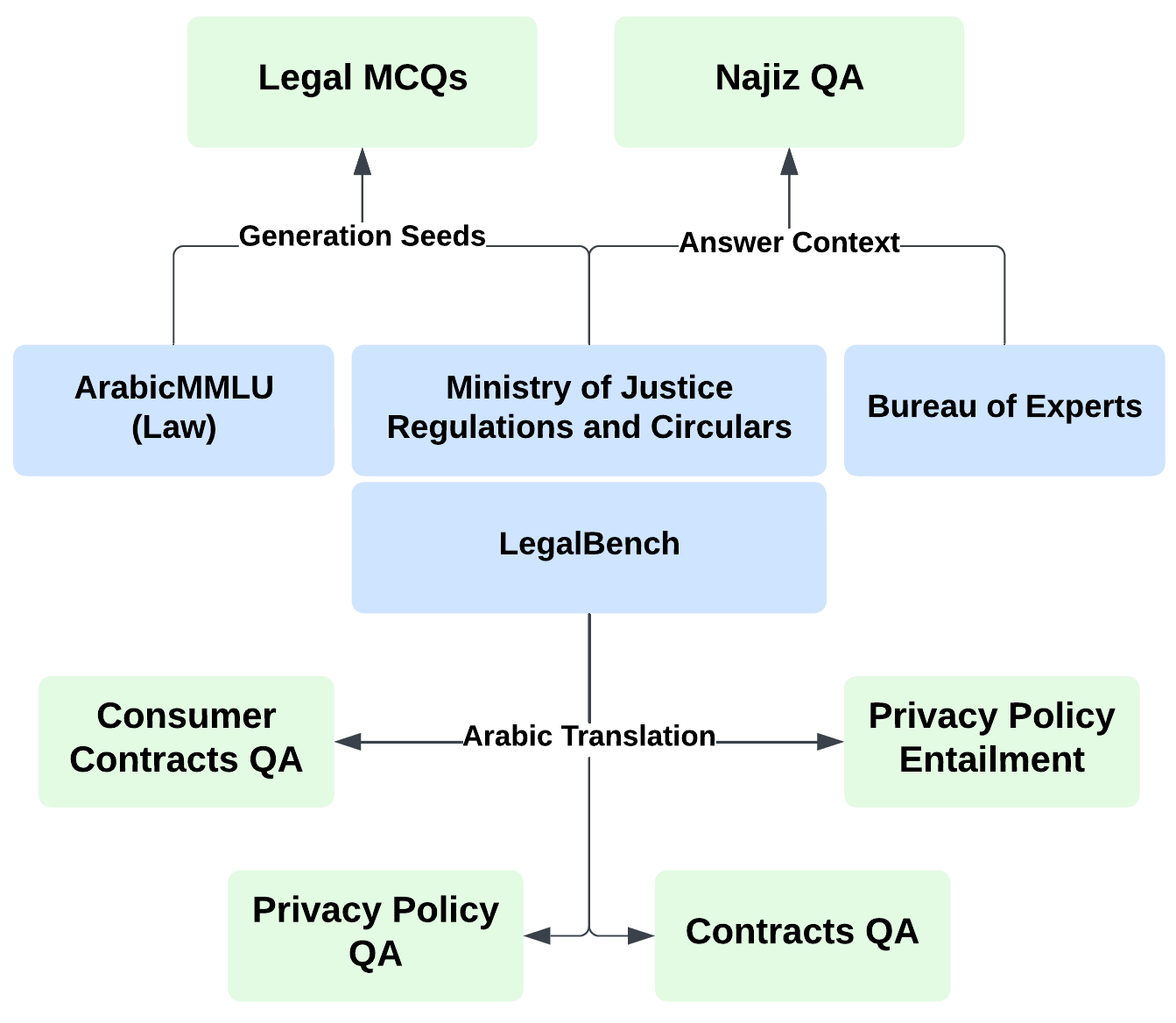}
    \caption{Tasks included in ArabLegalEval and their source documents.}
    \label{fig:data sources}
\end{figure}

% \begin{figure}[H]
%     \centering
%     \includegraphics[width=0.5\textwidth]{latex/sunburst_plot.png}
%     \caption{Tasks distribution in ArabLegalEval and their source documents.}
%     \label{fig:sunburstchart}
% \end{figure}

There are perhaps two broad categories of evaluation criteria that are useful for assessing the performance of legal LLMs. The first category is the ability of a model to use specific regulations, facts, and data that are relevant to a particular conversation. This can be achieved by having the LLM memorize specific facts, presumably by finetuning it on particular legal corpora, or more conveniently by using a Retrieval-Augmented Generation (RAG) system to retrieve information relevant to the context at hand.  The second broad category of assessment criteria is related to the model's ability to exhibit logical reasoning, understand relationships between entities and events, and apply these skills to answer questions. 

In this initial release of \verb|ArabLegalEval|, we include tasks to assess the legal reasoning capabilities in Arabic as well as tasks to measure the ability to recall and use legal knowledge embedded in the finetuned models.
% we primarily focus on the legal reasoning capabilities and we also include tasks to measure the capabilities of recalling pre-existing knowledge in the model weights. 
Much like \verb|ArabicMMLU| \cite{ArabicMMLU} was designed to test the general reasoning capabilities of Arabic LLMs but includes a mix of tasks with some requiring previous knowledge, the \verb|ArabLegalEval| benchmark seeks to develop Arabic legal tasks and questions derived from original Arabic legal sources, in consultation with legal professionals, to test the legal reasoning capabilities of Arabic LLMs. The benchmark also includes some high quality translations, verified by legal experts, of tasks from English legal benchmarks (\verb|LegalBench|) \cite{guha23_legalbench}.

The benchmark includes Arabic legal Multiple Choice Questions (MCQs), Question \& Answer (QA) pairs where the relevant Saudi regulations are included in the question, as well as Arabic translations of tasks from \verb|LegalBench|. 
The benchmark does not assess the ability of models to retrieve specific facts and laws relevant to a particular context from external knowledge bases. While important for the successful deployment of a legal LLM, these retrieval abilities will be assessed in future tasks. 

The primary contributions of this paper are: the development of a novel methodology for generating a legal QA dataset that can be adapted to other domains, such as finance; and \verb|ArabLegalEval|, an initial dataset specifically designed for assessing Arabic legal knowledge in large language models.

% The rest of this paper is organized as follows. Section 2 briefly reviews related work. Section 3 describes the data sources we relied on for the development. Section 4 describes the tasks currently included in the benchmark including how they were created, evaluated, and refined. Section 5 shows the performance of Arabic-capable models on the benchmark. Section 6 concludes and  outlines future work.  

%% file: latex/relatedwork.tex
\section{Related work}

\subsection{Arabic and multilingual reasoning capabilities in LLMs}

The success of LLMs, such as \verb|GPT-4| and \verb|GPT-4o| \cite{openai2024gpt4, openai2024hello}, \verb|Claude-3| \cite{anthropic2024claude32}, \verb|Command R| and \verb|Command R Plus| \cite{cohere2024commandr, cohere2024commandrplus}, and \verb|Llama3| \cite{meta2024llama3}, in exhibiting general purpose reasoning abilities - when queried in English - have naturally led to the development of models trained on Arabic content as well as benchmarks to evaluate the quality of reasoning in Arabic. \verb|Jais| \cite{jais23},
% and \verb|AceGPT| \cite{huang2023acegpt}, to models trained on low resource languages like \verb|Aya-23B| \cite{aryabumi2024aya}
for example, trained on native Arabic and quality translations, has shown better performance than other open-weight models on an Arabic version of the \verb|MMLU| multiple-choice school exam questions benchmark \cite{ArabicMMLU}. 
Arabic content and questions are also part of many broadly used multilingual benchmarks for assessing general reasoning. For example, MLQA \cite{lewis20_mlqa} assesses reading comprehension and question answering capabilities in languages including Arabic. 

While substantial experimentation is ongoing for evaluating reasoning and knowledge-based tasks in multilingual models, there are some model scale effects that appear to be emerging. On one hand, smaller models and models trained with limited data do not generally perform very well. Even \verb|GPT-3.5| \cite{ouyang2022gpt35} performs worse when responding to queries in Arabic and Non-English languages as compared to English on various tasks \cite{ArabicMMLU,jin23_inenglish}. On the other hand, the performance generally improves steadily with increasing model scale \cite{shi23}. \verb|GPT-4|, for example, exhibits remarkable performance on all sections of \verb|ArabicMMLU| surpassing all other models including Arabic-centric ones, even when few-shot prompts are used. It appears that 
multilingual reasoning is an emergent ability of large language models. Beyond a certain (task-dependent) scale, LLMs have evidently strong multilingual reasoning abilities. 

% \subsection{Legal LLMs}

% \subsection{Arabic LLMs.}

\subsection{LLMs and evaluation benchmarks in legal domains}

Given the importance of natural language to law, the advancing capabilities of large language models have been very quickly recognized and used for performing legals tasks. Current efforts are ongoing to explore whether current LLMs can be used as legal assistants for producing background research, drafting initial documents, summarizing contracts, answering questions about reports, and handling related tasks \cite{nay23,perlman22,goth23}. The announcement that \verb|GPT-4| has ``passed the bar exam'' \cite{katz24} has shown potential that intelligent legal advisors are not too far off. In the area of tax law for example, LLMs, particularly when combined with prompting enhancements and the correct legal texts, can perform at high levels of accuracy \cite{nay23}. Systems for Arabic court rulings and QA related to Legal Palestinian cooperative maters were presented in \cite{ammar2024prediction, maree2024transforming}.

However, closer inspection has revealed that certification exams or narrow domains are not always representative of the actual use-cases in law practice or for LLMs \cite{fei23_lawbench}. As a result, a number of efforts have been ongoing to develop models, agents, and benchmarks that use tasks and 
general conceptual frameworks similar to those used in the legal profession. \verb|LegalBench| and a Chinese counterpart, \verb|LawBench| \cite{fei23_lawbench}, have sought to collect and categorize a broad set of tasks that appear in law practice, from simple tasks requiring  rule-recall or issue recognition to sophisticated tasks that require interpretation, drawing conclusions, and multi-hop reasoning.  As many of the LegalBench tasks, particularly those requiring interpretation, appear to involve reasoning skills that are not particular to US law, it seems reasonable to use translated versions of these tasks for testing LLMs in different languages.

% consumer_contracts_qa, which evaluates an LLM’s ability to determine the rights/obligations imposed by terms of
% service clauses from popular websites (Section F.5).
% contract_qa,whichevaluatesanLLM’sabilitytoidentifydifferenttypesofcontractualprovisions.

% privacy_policy_entailment [161], which evaluates an LLM’s ability to answer entailment questions from privacy policies (Section F.22).
% • privacy_policy_qa[112],whichevaluatesanLLM’sabilitytodetermineifaclausefromaprivacypolicycontainsthe answer to a particular question (Section F.23).

% - related work in Legal in LegalBench, other languages: LawBench
% - Arabic Legal
%     - 
% - Arabic benchmarks:
%     - MMLU
%     - ArabicMMLU
% - domain specific LLMs

%% file: latex/arabLegalEval.tex
\section{Data Sources}

% \begin{itemize}
%     \item from native arabic data, Najiz
%     \item from MOJ, all with context
%     \item Najiz mostly without without context 
% 	 \item with context
% 	 \item evaluated with LLM on top
% \end{itemize}

% \begin{table*}[h]
% \centering

% \begin{tabular}{|l |l |l |l |l |l |l|} \hline 
% source & task & \# & synthetic & human signal & \% quality assurance & propagated error rate \% \\ \hline 
% NajizFAQ & Open ended QA + Context & 490 & - & human filtered?? &  &  \\ \hline 
% LegalBench contract\_qa & ClosedQA + Context & ??? & machine translated & translation 100\% approved by human expert &  &  \\ \hline 
% MOJ & MCQ & ??? & used as source documents to generate MCQ & human expert filtered + AI filtered &  & 1 - (0.95 * ) \\ \hline 
% Arabic MMLU (Legal and Political science subsets) &  & ??? & - &  &  &  \\ \hline

% \end{tabular}
% \end{table*}

% \subsection{Sources of Legal Data}

\subsection{Raw Arabic data sources}

Obtaining comprehensive Arabic legal data is challenging. After consulting Saudi legal experts, we identified key sources that we have started to incorporate (Figure \ref{fig:data sources}). These include scanned documents from the Ministry of Justice (MoJ, \<وزارة العدل>) and the Board of Experts (\href{https://laws.boe.gov.sa/BoeLaws/Laws/}{BoE}, \<هيئة الخبراء بمجلس الوزراء>). MoJ documents cover 67 regulations (5720 subjects) and 388 circulars, while BoE documents include 448 rules (14134 subjects). See Appendix \ref{sec:Examples of Data Sources} for samples.

Both sets of documents contain regulations and statutes, they differ primarily in the topics they cover. MoJ documents specialize in topics issued by the Ministry, providing a comprehensive database of judicial regulations and legislation essential for legal research and practice.

Note that all data sources used are open and publicly available and scraped from official websites with no confidential information. These were the raw data sources used to generated the benchmark

% start of faris's section

% Scraped and filtered human-generated data from frequently asked
% questions related to the Ministry of Justice and the Board of Experts in Saudi Arabia.

  % In addition 
%   \tightlist
%   \item
%     We select generalizable data that is not specific to any country,
%     making it logical to translate and expect an Arabic model to solve.
%     We focus on contract-related questions, providing the full context
%     and replicating the experimental setup used in LegalBench.
%   \end{itemize}
% \end{itemize}

% \begin{itemize}
%     \item This serves as the main source of legislation in the Kingdom of Saudi Arabia. The \href{https://laws.boe.gov.sa/BoeLaws/Laws/}{BOE} website provides access to a comprehensive collection of legal documents.
%     \item \textbf{MOJ Regulation} This source encompasses judicial regulations and legislation, providing a comprehensive database of
% legal norms essential for legal research and practice.

%     BOE is used for context in NajizQA dataset.
    
% \end{itemize}

% \textbf{BOE (Board of Experts at the Council of Ministers)}

\subsubsection{Preparation Steps}

For both BoE and MoJ documents, we followed a systematic data preparation process to ensure that the data is rich and easy to work with. The data was scraped from the web to capture all the regulations while preserving all metadata to allow for careful filtering later.

% \vspace*{-4pt}
%     \begin{itemize}
%         \setlength{\itemsep}{-2pt}
%         \item Data Extraction: The data was extracted using a custom web scraping script, which captures all the regulations.
%         \item Organizing Data: The extracted data was organized into a structured database format to facilitate analysis.
%         \item Metadata Preservation: All metadata related to the documents and their provenance are kept to enhance data richness and usability.
%     \end{itemize}

An example of a processed, structured, MoJ data document is shown in Figure 
\ref{fig:moj-example}. 

% \ref{fig:boe-example} 

\subsubsection{Frequently asked Questions (FAQs)}

In addition to these raw sources, we also rely on human-written FAQs that are publicly available. The questions and answers in this data are generally available in the BoE and MoJ documents, and we use them to build an open-ended question answering task in the benchmark, which we call \textit{NajizQA}. A sample of this data can be found in Figure \ref{fig:najizqa-example}.

\subsection{LegalBench}

In addition to native Arabic sources, we rely on the translation of English legal documents. 
\verb|Legalbench| is a benchmark for legal reasoning in English LLMs \cite{guha23_legalbench}. We selected four datasets from it and translated them from English to Arabic. 
% The datasets include ``Consumer Contracts QA'', ``Contracts QA'', ``Privacy Policy QA'', and ``Privacy Policy Entailment''. 
These were specifically chosen because they have fewer localization requirements and are more universal. This makes them ideal for assessing the ability of LLMs to understand and interpret legal clauses and contracts in Arabic.\\
% \begin{enumerate}
\textbf{\textit{Consumer Contracts QA:}} this dataset consists of 400 yes/no questions about the rights and obligations outlined in online terms of service agreements. \\
% [We need to add reference here Berkeley Technology Law Journal ]
\textbf{\textit{Contracts QA:}} this dataset consists contract clauses and questions about these contracts. It has 88 examples, with 80 examples for testing and 8 examples for training.\\
\textbf{\textit{Privacy Policy QA:}} the dataset consists of questions and corresponding clauses from a privacy policy. It consists of a total of 10,931 instances, with 8 examples for training and 10,923 for testing. \\
\textbf{\textit{Privacy Policy Entailment:}} this dataset has 4385 examples in training and testing, each example is a privacy policy clause and a description. The goal is to determine if the description for the clause is correct or not. 
% \end{enumerate}

\subsection{ArabicMMLU}
\verb|ArabicMMLU| \cite{ArabicMMLU}, an Arabic knowledge evaluation benchmark constructed from human-written school exams from Arabic-speaking countries, served as one of the inspirations for this work. With a subset of its samples focused on the legal domain, \verb|ArabicMMLU| provided a valuable starting point for us to generate our MCQs.

%% file: latex/MCQ.tex
% \begin{figure*}[h]
%     \centering
%     \includegraphics[width=\linewidth]{latex/assets/MCQs_gen_filter.png}
%     \caption{MCQs Generation and Filtering}
% \end{figure*}

\subsection{MCQs}

% Having scraped a lot of legal documents from the Saudi MOJ and BOE as mentioned before. We want to turn these valuable resources into a useful benchmark that can be used to evaluate LLMs’ knowledge and ability to reason about, and comprehend text and language form the legal domain relating to Arabic and Saudi culture. To achieve that, we utilized these documents as seeds for large language models (LLMs) to generate multiple-choice questions (MCQs). 

One standard method of benchmarking reasoning and memorization capabilities in neural networks are MCQs, such as \verb|MMLU| \cite{MMLU}. It is easy to verify the correctness of the answer using exact matching or regular expressions making MCQs ideal for automatic evaluation.

% Even though MCQ benchmarks may not be robust, for example it could be very sensitive to perter

\begin{figure}[h]
    \centering
    \vspace*{-4pt}
    \includegraphics[width=6.8cm]{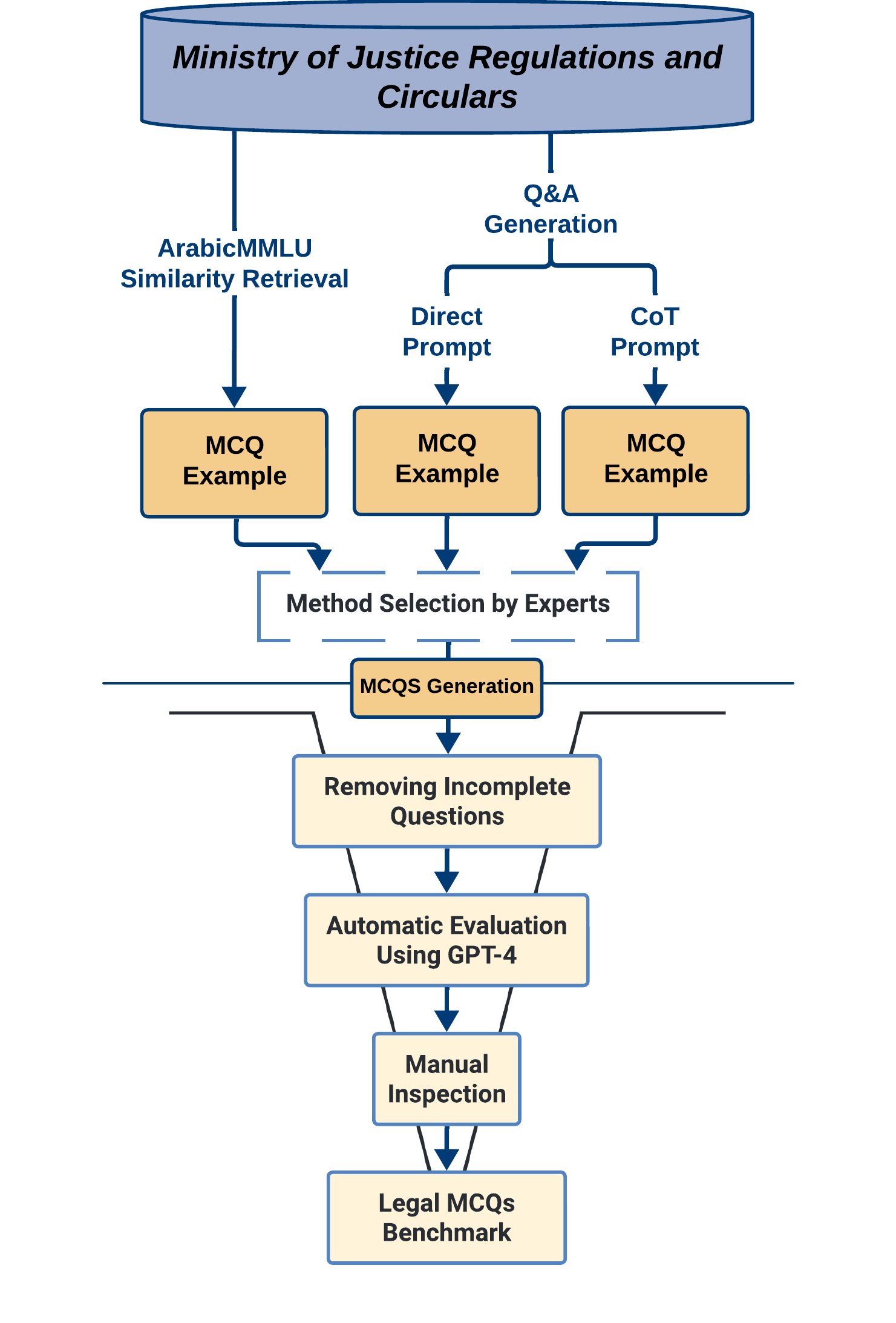}
    \caption{MCQs Generation and Filtering}
    \label{fig:MCQs_generation}
\end{figure}

% MCQs are preferred for evaluating understanding due to ease of assessment.
Given the availability of a large collection of raw legal documents from the MoJ and BoE, we aim to generate synthetic MCQs from them, using them the documents as context.
Generating MCQs poses two main challenges: formulating questions and generating options (correct answers and plausible distractors).\\
We approached this using a robust LLM and experimented with three methods: 1. \textit{QA to MCQ}, 2. \textit{Chain of Thought (CoT)}, and 3. \textit{Retrieval-based in-context learning}. See Figure \ref{fig:MCQs_generation}.

In all cases, we prompt the model to synthesize questions in the same format and style as \verb|MMLU|.

\subsubsection{QA to MCQ}
% As our baseline, we experimented with LLMs' ability to generate MCQs through a
Here we use a two-step prompt. Given a legal document, the model is prompted to generate both a question and its answer, then a follow-up prompt to convert the question into a Multiple Choice Question (MCQ) by rephrasing the answer and generating appropriate distractors,
\vspace*{-6pt}
\[
f(c_i) \rightarrow q_i, a_i; \quad 
% \] 
% \[
g(q_i, a_i) \rightarrow D_i
\]
where $f$ is a language model that is given some legal context $c_i$, and then prompted to generate a QA pair $(q_i, a_i)$, and $g$ is another instance of the model that is given the QA pair and is tasked to generate a set of distractors $D_i$ for the question.
\subsubsection{QA to MCQ with CoT}
% One key idea behind MCQs’ distractors is the they shouldn’t be completely  random, instead they should be based on reasoning to avoid being obviously incorrect.
CoT is a relatively recent method of prompting LLMs where instead of directly producing the answer, the model is given space to reason and “think out loud” before answering, essentially providing itself with more context.
This simple technique has led to relatively huge gains in performance, in a variety of tasks \cite{wei2023chainofthought}.
We want to utilize this idea to generate reasoning-based plausible distractors. This approach can be formulated as:
\vspace*{-4pt}
\[
f(c_i) \rightarrow q_i a_i; \quad
% \]
% \[
g_{\text{CoT}}(q_i, a_i) \rightarrow D_i
\]
where $f$ is a language model prompted with some context $c$ from the scraped MoJ documents to produce a question $q$ and its answer $a$, these are then fed to another instance of the model $g$ with CoT prompt to produce a set of plausible distractors $D$ for the question.

\subsubsection{Direct MCQs generation with in-context examples}

Since the inspirations of this work are popular knowledge evaluation benchmarks such as \verb|MMLU| and \verb|ArabicMMLU|, our goal is to generate questions in a similar format. We begin by taking a subset of \verb|ArabicMMLU| questions that have ‘Law’ as their subject tag (about 300 examples).  For each document in the MoJ dataset, we perform a semantic similarity search \cite{risch-etal-2021-semantic} to retrieve the top $k$ examples from these questions. These examples are then added to the prompt to guide the model in generating questions of a similar style.
\vspace*{-4pt}
\[
f(c_i, E_k) \rightarrow q_i, a_i, D_i
\]

The MoJ context is augmented with a set of \verb|ArabicMMLU| examples $E_k$ that are fed to a model $f$ to generate a question $q$, where $k$ is the number of retrieved examples. These examples provide context for the model for in-context learning so that the generated answer has the same style and format as the \verb|ArabicMMLU| examples. From this, the model generated a question $a$ and all of the distractors $D$ in one go.

\subsubsection{MCQ filtering and curation}

Each of the above techniques was tested and then had the results manually reviewed and inspected by the legal experts according to the metrics in Appendix \ref{mcq_evaluation_criteria}

% \begin{itemize}[itemsep=1pt]
%     \item Is answer correct?
%     \item Is there only one right answer?
%     \item Is the other options related to the question?
%     \item Is the Context needed to answer the question?
%     \item Is the question relevance to the Context?
%     \item (From 0-5) Rate the Answer relevance to the Context
% \end{itemize}

It was concluded that the best method for generating MCQs was \textit{in-context examples} with $k=3$. 
Based on this, we decided to use this technique to generate all the MCQs.

After generating the MCQs dataset (approximately 12k samples), we did some automatic filtering using \verb|GPT-4| \cite{chiang2023large}, where it was prompted to check if each sample satisfied our criteria (see appendix \autoref{mcq_evaluation_criteria}). Any sample that failed to satisfy all of the above were removed, which left us with 10,583 MCQs for our benchmark. At the end, we extracted a random subset of the dataset for a final manual inspection.

% we used Yamane’s formula \cite{yamane67} to extract a random subset of the dataset for a final manual inspection.
% , to calculate a potential maximum score on the benchmark and identify any existing issues with the generated samples.

% Any sample that failed to satisfy all of the above were removed, afterwards, we used Yamane’s formula to extract a subset of the dataset for manual inspection, to calculate a potential maximum score on the benchmark and identify any existing issues with the generated samples.

% \begin{equation}
% n = \frac{N}{1 + N(e^2)}
% \end{equation}
% \captionsetup{labelformat=empty}
% \caption{equation}{Yamane's formula}

% With {N = 10583} and {e = 0.05}, we calculate a sample size of {n = 385} to manually inspect.

% \textbf{discuss results when done}

\subsubsection{Models for Generation}
\begin{figure}[H]
    \centering
    \includegraphics[width=1\linewidth]{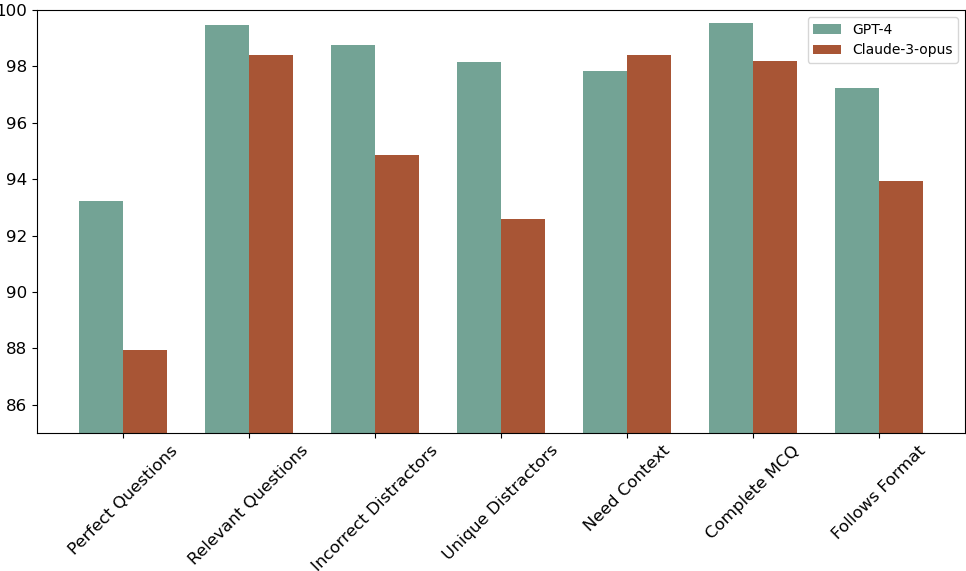}
    \caption{GPT-4 vs Claude-3-opus MCQs Generation}
\end{figure}

It has been observed that models tend to perform better on synthetic data generated by themselves as apposed to another model \cite{ft-judge-models}.
To mitigate this unfair advantage, we split our documents and alternate between two state of the art models: \verb|Claude-3-opus| and \verb|GPT-4|. We make sure a model isn't evaluated on questions generated by itself.

% We split our source documents (around 6,000 documents) into two halves. We used GPT-4 with one half and Claude-3-opus with the other, prompting each model to generate two questions per document. Each of these two models was then evaluated using the half of the dataset generated by the other. For the remaining models, both halves were merged into one dataset for evaluation.

%% file: latex/QA.tex
\subsection{QA}
This dataset includes
filtered legal QAs from a publicly available human written set of FAQs. The questions include  only those referencing specific statues and regulations and articles in the MoJ and BoE Arabic documents, and are therefore particularly valuable for evaluating Arabic legal LLMs. See figure \ref{fig:NajizQAproc}. A sample of the data shown in the Appendix \ref{sec:Examples of Data Sources}.  

Embedding techniques were used in the semantic similarity matching phase (Figure \ref{fig:NajizQAproc}). Specifically, we utilized \verb|text-embedding-3-small| \cite{openai2024embedding} to generate embeddings for both the questions and contextual information. Subsequently, \verb|cosine similarity| was employed to identify relevant texts corresponding to each question. This methodological framework was selected to evaluate the models' reasoning capabilities with and without context.
\begin{figure}[H]
    \centering
    \includegraphics[width=1\linewidth]{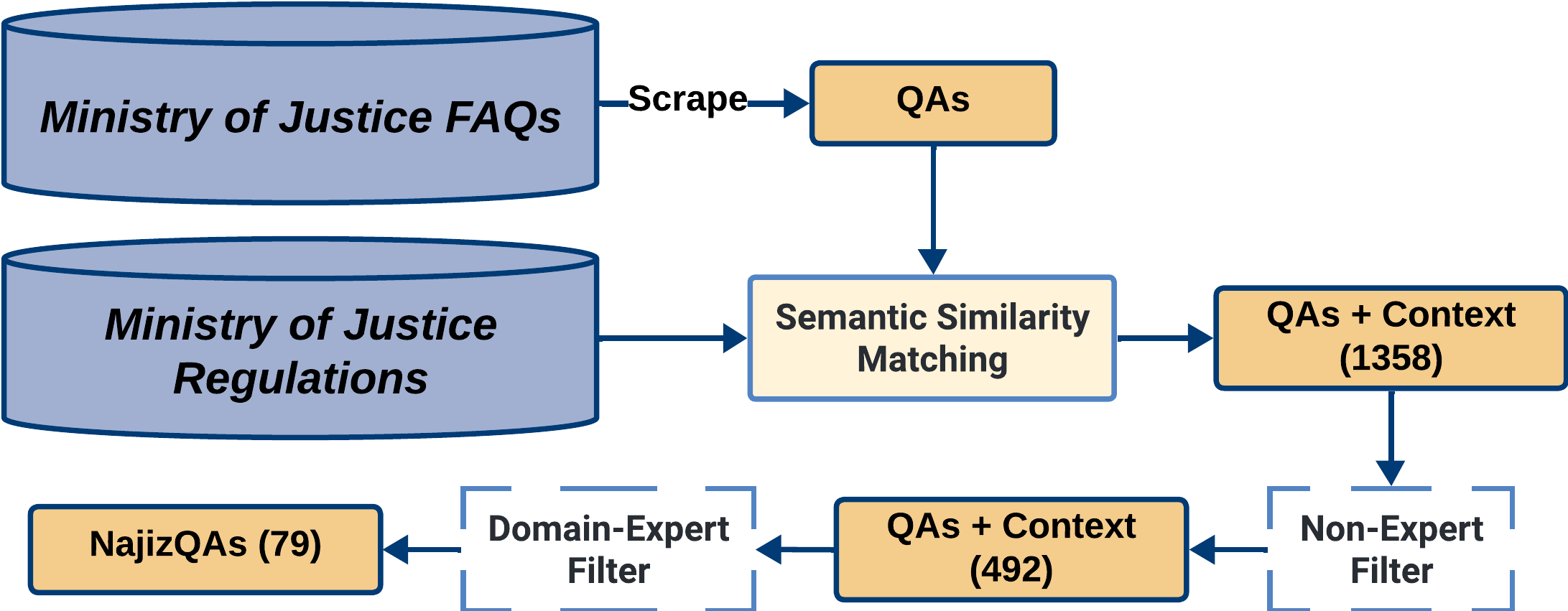}
    \caption{NajizQA curation pipeline}
    \label{fig:NajizQA_proc}
\end{figure}

% The preparation of these questions involved:
% \vspace*{-6pt}
% \begin{itemize}
%     \setlength\itemsep{-3pt}
%     \item Extracting embeddings for the questions as well as for the sources collected from MOJ using the OpenAI's \verb|text-embedding-3-small| model to find the context for the questions.
%     \item Initial threshold setting: A non-legal-expert reviewed the data and set a threshold of 0.6 based on their review.
%     \item Filtering: The initial threshold reduced the number of questions from 1358 to 492.
%     \item Sampling: A sample of 7.5\% (65 questions) of the filtered questions was taken.
%     \item Expert Review: A legal expert reviewed the sample and suggested setting the threshold to 0.72, resulting in a final set of 79 questions.
% \end{itemize}

% \begin{figure}[h]
%     \centering
%     \includegraphics[width=0.5\textwidth]{latex/NajizQA-examples-table.png}
%     \caption{Najiz Question Answering dataset with context}
%     \label{fig:najiz-example}
% \end{figure}

%% file: latex/arlegalbench.tex
\subsection{Arabic translation of LegalBench}

\subsubsection{Translation Strategy}
% Identifying the most suitable model is a critical step in obtaining a high-quality translated dataset. In our study, 

We evaluated three different machine translation models, Azure Translation Services \cite{AzureTranslateAPI}, Google API Translation \cite{GoogleTranslateAPI}, and Opus MT \cite{TiedemannThottingal:EAMT2020} along with \verb|GPT-4| to determine the best translator for a publicly available dataset with legal context. The dataset used in these experiments is the \textbf{United Nations Parallel Corpus} \cite{ziemski16}, which consists of UN documents in the six official UN languages.
% Arabic, Chinese, English, French, Russian, and Spanish. For the purpose of our research,
We focused solely on the English and Arabic datasets, which comprise 20 million rows.
% The English and Arabic dataset in the UN Parallel Corpus comprises 20 million rows. 
However, to expedite the experimentation process, we utilized a subset of 14,000 examples from this dataset. This subset was used to evaluate the performance of the selected models without any preprocessing. Rouge metrics were used in evaluating the translation quality.

% This choice was motivated by the unique challenges Arabic poses for machine translation especially in the legal domain. By using a dataset representative of real-world legal text, we can accurately assess the model’s performance in producing high-quality translations. 
% The evaluation process involved assessing each model's translation quality based on Rouge metrics, Rouge stands for ``Recall-Oriented Understudy for Gisting Evaluation'' and it is a set of metrics used to measure the performance of machine generated text and commonly used in translation and summarization. Rouge score assesses the similarity between the machine-generated text and a reference text and is calculated by comparing the overlap of n-grams between the machine-generated text and the reference text. The score ranges from 0 to 1 with 1 indicating higher degree of similarity between texts.

% In our evaluation, we employed three ROUGE metrics: ROUGE-1, ROUGE-2, and ROUGE-L, to assess the quality of the machine-generated translations against the reference translations. ROUGE-1 measures the overlap of unigrams (single words), providing a basic understanding of lexical similarity. ROUGE-2 measures the overlap of bigrams (two-word sequences), offering insights into contextual coherence. ROUGE-L measures the longest common subsequence (LCS), evaluating sentence-level structure similarity and fluency.

By comparing the results obtained from the different models, we aimed to identify the one that consistently produced the best translation from English to Arabic in the legal domain. The results of our experiments on this subset of data, without any prepossessing, are shown in Table \ref{tab:rouge_r} in Appendix \ref{sec:translation}. 
Overall, the Opus model outperformed the other models across all the metrics. The Opus model achieved a ROUGE-1 score of 0.52, a ROUGE-2 score of 0.3, and a ROUGE-L score of 0.51. In addition to this, we had a sample of the translations manually reviewed by legal experts.

% We observed that Rouge-1 scores were consistently higher across all the models followed by the Rouge-L scores, and this indicates that the models performed better in capturing unigram overlap and the longest common subsequence between the machine-generated translations and the reference translations.  Overall, Opus model outperformed the other models across all the metrics. The Opus model achieved a ROUGE-1 score of 0.52, a ROUGE-2 score of 0.3, and a ROUGE-L score of 0.51.

These findings established a basis for selecting Opus as the translator for our main task: translating selected datasets from \verb|LegalBench| from English to Arabic. 

\subsubsection{Evaluation of Translated Output}
To ensure the highest translation quality, we employed the three models from our initial experiments to translate the ``Consumer Contract QA'' dataset. We conducted a manual inspection of the overall results and, to further enhance the quality assessment, engaged both \verb|GPT-4| and legal experts as evaluators for the translated content.

\paragraph{GPT-4 Evaluation}
We tasked \verb|GPT-4| with evaluating a sample of the translated text alongside the original text, asking it to rate the translation quality on a scale from 1 to 5. Interestingly, \verb|GPT-4| consistently rated the translations as either 4 or 5 out of 5, indicating a high level of perceived quality, see Appendix \ref{sec:translation} for \verb|GPT-4| evaluation prompt and result example.
% Examples of the evaluation: 

\paragraph{Human Evaluation}
We selected five examples from the translated texts, each consisting of a contract text and a corresponding question. To evaluate the performance of the three translation models — Azure Translation Services, Google API Services, and Opus MT — we asked legal experts to assess each contract and question separately. The experts provided scores for each model's based on the translation quality and how well the legal context was preserved in the translation.
The overall scores for all three models were relatively close, but Opus MT consistently achieved the highest score among them:
\vspace{-4pt}
\begin{itemize}
\setlength\itemsep{-4pt}
\item Google API: average score of 3.3 out of 5
\item Azure: average score of 3.6 out of 5
\item Opus MT: Average score of 4 out of 5
\end{itemize}
In addition to the overall scores, we asked the experts to rank the best model for each example. Opus MT was chosen as the best model 60\% of the time, while Azure Translation Services was selected 40\% of the time. Interestingly, Google API Translation was never selected as the best model for any of the examples. 

These results suggest that while all three models performed reasonably well, Opus MT demonstrated superior translation quality for the given legal texts and questions, as determined by the expert evaluations.
See Figure \ref{fig:sample_transated_Data} in Appendix \ref{sec:translation} for a translated example.

%% file: latex/mcq_eval.tex
%--------------------------------REEM-------------------------------------
\subsection{MCQ Evaluation}
In this section, we evaluate the performance of language models on our synthetic MCQs dataset using tailored prompts, where the instructions in the prompt are provided in English for each model.

\subsubsection{Experiment Setup}
We aim to improve the models’ capabilities by modifying the given prompt. Different parts of the prompt can be optimized according to a given metric, and in this evaluation, we started with optimizing the instruction and few-shot examples to determine which method is more effective. Unfortunately, instruction optimization yielded no significant performance gain. On the other hand, few-shot optimization boosted the performance of many models. Hence, we decided to focus on few-shot optimization. \verb|ArabicMMLU| is a benchmark to assess the capabilities of models, similar to \verb|MMLU| benchmark, but with localized data in Arabic. A subset relevant to the legal domain of \verb|ArabicMMLU| was sampled resulting in a total size of 524 questions after filtering questions that require context. Out of those 524 samples, 314 and 210 are law and political science, respectively. We optimize the prompts on this subset to use it for evaluating our generated MCQs.

\begin{figure}[h]
    \centering
    \includegraphics[width=\columnwidth]{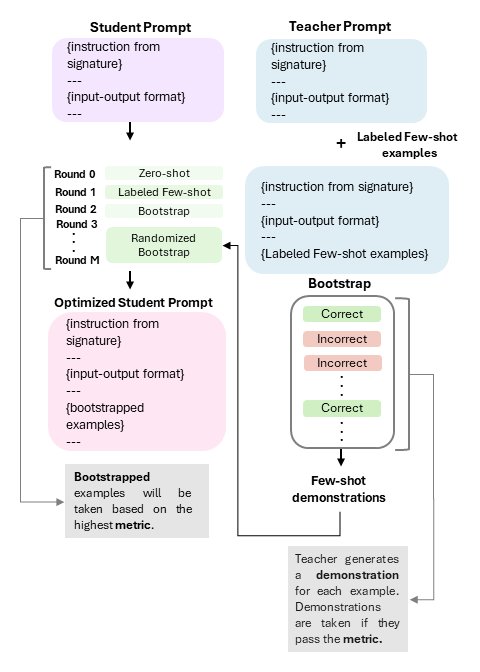}
    \caption{DSPy's prompt optimizer process.}
    \label{fig:bootstrap.png}
\end{figure}

DSPy \cite{DSPy} is a Language Model (LM) programming framework to optimize LM prompts and weights automatically by recompiling the entire pipeline to optimize it on a specific task. We relied on this framework for prompt optimization. Initially, all of the models were given a zero-shot prompt with an English answer instruction and the input-output format. Then, this zero-shot prompt is optimized for each model to achieve a higher performance by augmenting it with either plain few-shot examples or few-shot with reasoning demonstrations using CoT. Teacher and student models were used to create few-shot examples with CoT demonstrations, where the teacher is either a clone of the student or another model. Figure \ref{fig:bootstrap.png} demonstrates DSPy's bootstrapped few-shot optimization.

\vspace*{-6pt}
\begin{table*}[ht]
\centering
\caption{Experimental results on Generated MCQs}
\label{tab:ex_genmcq}
\scalebox{0.85}{
\begin{tabular}{cccccc}
\hline
\multicolumn{1}{|c|}{}           & \multicolumn{1}{c|}{Original} & \multicolumn{1}{c|}{Few-shot}         & \multicolumn{1}{c|}{Few-shot (GPT-4 Teacher)} & \multicolumn{1}{c|}{CoT Few-shot} & \multicolumn{1}{c|}{CoT Few-shot (GPT-4 Teacher)} \\ \hline
\multicolumn{1}{|c|}{GPT-4o}     & \multicolumn{1}{c|}{76.80\%}  & \multicolumn{1}{c|}{\textbf{79.10\%}} & \multicolumn{1}{c|}{-}                     & \multicolumn{1}{c|}{76.50\%}      & \multicolumn{1}{c|}{-}                         \\ \hline
\multicolumn{1}{|c|}{Command R}  & \multicolumn{1}{c|}{68.10\%}  & \multicolumn{1}{c|}{71.40\%}          & \multicolumn{1}{c|}{72.30\%}                  & \multicolumn{1}{c|}{71.00\%}      & \multicolumn{1}{c|}{70.09\%}                      \\ \hline
\multicolumn{1}{|c|}{Command R+} & \multicolumn{1}{c|}{68.10\%}  & \multicolumn{1}{c|}{71.40\%}          & \multicolumn{1}{c|}{72.30\%}                  & \multicolumn{1}{c|}{73.10\%}      & \multicolumn{1}{c|}{69.86\%}                      \\ \hline
\multicolumn{1}{|c|}{Llama3 (8B)}  & \multicolumn{1}{c|}{68.20\%}  & \multicolumn{1}{c|}{71.20\%}          & \multicolumn{1}{c|}{72.30\%}                  & \multicolumn{1}{c|}{73.60\%}      & \multicolumn{1}{c|}{70.45\%}                      \\ \hline
\multicolumn{1}{|c|}{Llama3 (70B)} & \multicolumn{1}{c|}{68.17\%}  & \multicolumn{1}{c|}{71.47\%}          & \multicolumn{1}{c|}{72.34\%}                  & \multicolumn{1}{c|}{71.93\%}      & \multicolumn{1}{c|}{70.53\%}                      \\ \hline
\multicolumn{1}{l}{}             & \multicolumn{1}{l}{}          & \multicolumn{1}{l}{}                  & \multicolumn{1}{l}{}                          & \multicolumn{1}{l}{}              & \multicolumn{1}{l}{}                             
\end{tabular}
}
\end{table*}
\subsubsection{MCQ Results Analysis}
The MCQ dataset contains 10,583 questions, with 5,544 generated by \verb|GPT-4| and 5,039 by \verb|Claude-3-opus|. We evaluated several LMs on this dataset, including \verb|Command R|, \verb|Command R Plus|, and \verb|Llama3 (8B and 70B)|. However, \verb|GPT-4o| was tested only on \verb|Claude-3-opus| subset to mitigate potential bias towards its own generated questions \cite{panickssery2024llmevaluators}. Our evaluation metric assesses LM performance by comparing the selected answer with the correct one for each question. For further \verb|ArabicMMLU| detailed results with optimized prompts, refer to appendix \ref{sec:arabicmmlu_opt}.

Table \ref{tab:ex_genmcq} summarizes the performance of LMs on our generated MCQs using prompts optimized with DSPy on \verb|ArabicMMLU|'s legal subsets. Interestingly, many of the optimized few-shot prompts shared identical examples, suggesting that certain examples play a more significant role in improving LMs’ performance than others. In addition, few-shot examples coupled with CoT reasoning boosted the capabilities of the models. For further testing, we employed \verb|GPT-4| as a teacher model for smaller LMs in both plain and CoT few-shot prompting. Surprisingly, these smaller LMs demonstrated greater performance in few-shot CoT when the teacher was a clone of themselves, rather than the more advanced \verb|GPT-4|. This unexpected result suggests that LMs may have a better grasp of their own reasoning. 

\begin{figure}[H] 
    \centering
    \includegraphics[width=\linewidth]{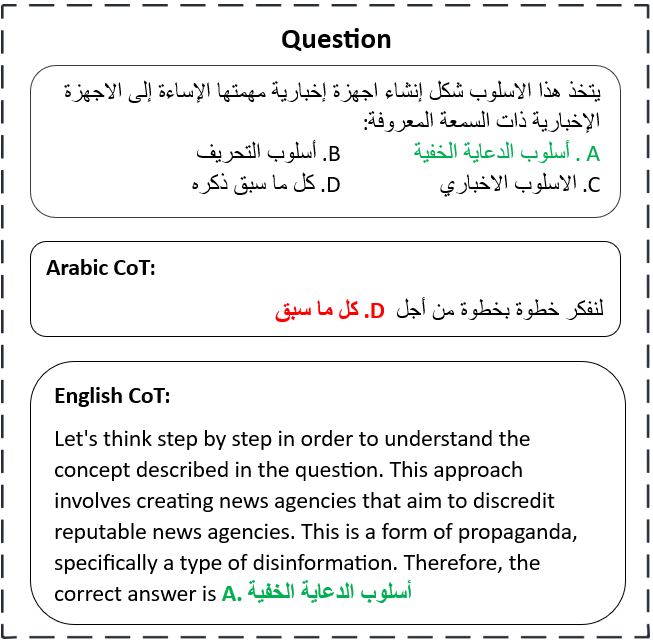}
    \caption{Command R Plus CoT Reasoning. The English translation is \textit{'This style takes the form of creating news networks whose purpose is to harm reputable news networks:
    A. The style of hidden propaganda
    B. The style of manipulation
    C. The journalistic style
    D. All of the above'}. The correct answer, A, is highlighted in green.}
    \label{fig:cot_reasoning.png}
\end{figure}

We observed that the choice of language plays a crucial role in the reasoning abilities of smaller models. In many cases, these LMs generated answers without providing accompanying reasoning. Figure \ref{fig:cot_reasoning.png} shows the differences in reasoning language for \verb|Command R Plus|, revealing a degradation of the model’s reasoning capability when the language choice is Arabic. 

Table \ref{tab:ex_genmcq} shows that \verb|GPT-4o| demonstrated superior performance across all prompting methods, achieving 79.10\% accuracy with few-shot prompting. \verb|Llama3 (8B)| achieved a 5\% increase using few-shot prompting with CoT reasoning. Similarly, the other LMs, except \verb|Command R|, obtained better results when the prompt included their own CoT reasoning. Our findings reveal potential for improving LMs' question-answering capabilities through task-specific optimized prompts.

%% file: latex/qa_eval.tex
% \subsection{Results}

\subsection{Open ended QA Evaluation}

% For open ended question answering, we use the NajizQA data.

\subsubsection{Experiment Setup}

% We categorize the Question~Answering (QA) task into two types: 1. Closed-set QA: where the answers are between several classes like “yes” and “no”, similar to classification. This is easy to evaluate using regular expressions.
% Contrary to the first type of QA: 
In open-ended QA---contrary to closed-set QA where the possible answers are fixed---the answer can be expressed in natural language, which is harder to evaluate since it relies on semantics and meaning. We observe that English prompts for the exam-taker LLM and the judge-LLM outperform Arabic prompts, so we use those English prompts for the instructions.

Traditional evaluation of QA models uses metrics like Exact Match (EM), F1-score, and top-n-accuracy, focusing on lexical matching. These metrics often miss semantically similar answers that use different words.

We use use an \textit{LLM-as-a-judge} \cite{ft-judge-models}, \cite{prometheus},  \cite{Zheng-et-al-judging} to rate the answer similarity on a scale from 0 to 5 given the generated answer and the reference ground truth answer.
In our case, \verb|GPT-4| is the judge, and we refer to the output score as the \textit{answer similarity metric}, see figure \ref{fig:llm-as-a-judge.png} for details. We notice that the judge LLMs give a lower score to answers that are in a different language than that of the reference answer even when the content is correct. To mitigate this, we prompt the models to output answers in the same language as the reference answer (which is Arabic in our dataset).

We run the evaluation for each LLM with two cases: one where it is given the question and the needed context, and one where it is given only the question.
% TODO: put example question here and there
We also add a "golden model" that always answers the perfect ground truth answer, just so that we can compare against the upper bound of what the judge-LLM is going to score.

% \textbf{CHANGE THE LIST TO AN IMAGE LATER}

\subsubsection{QA Results}

We run our experiments on the 79 NajizQA pairs that have been filtered and verified by legal experts.
% TODO: reference to the NajizQA sample table
% TODO: move the NajizQA sample table to the appendix

\begin{figure}[h]
    \centering
    \includegraphics[width=1\columnwidth]{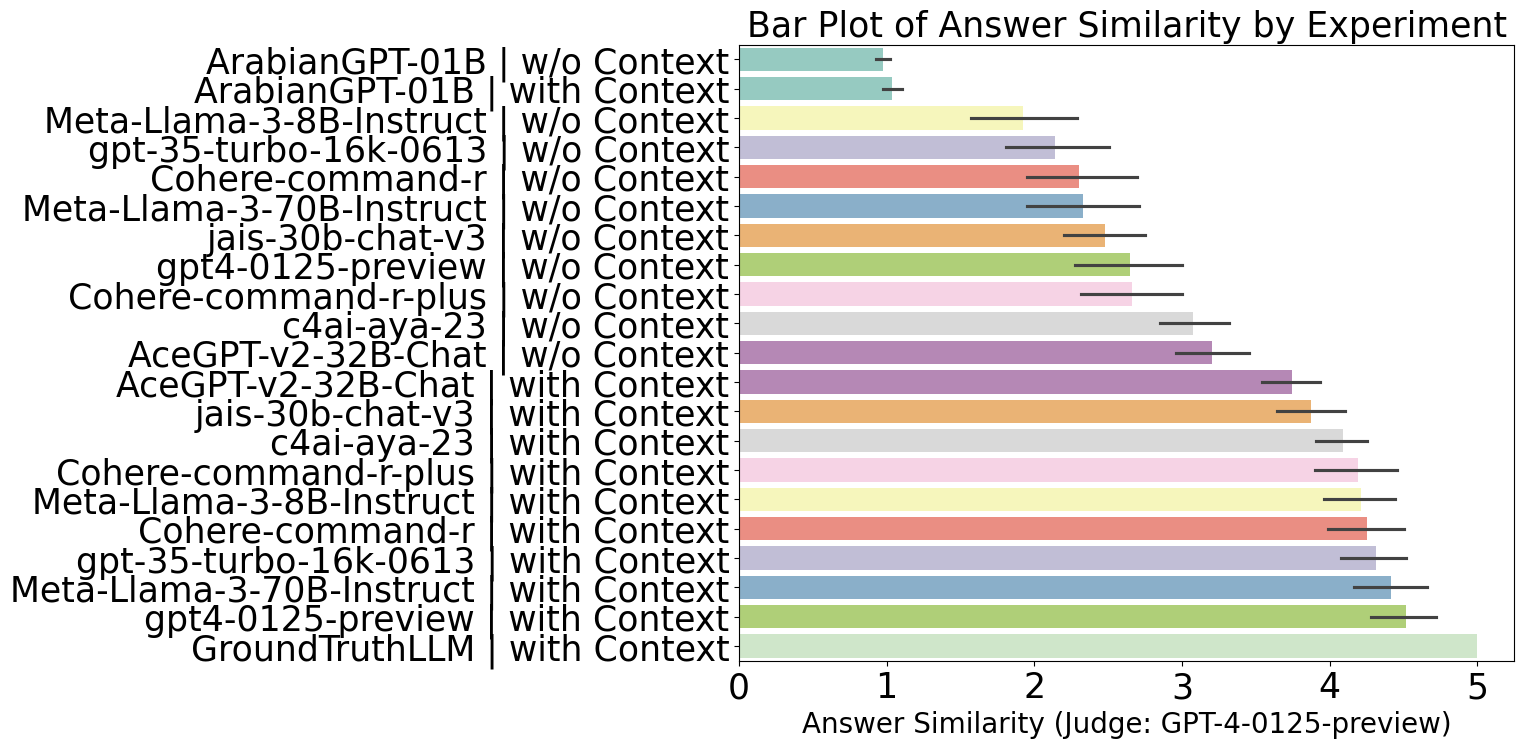}
    \caption{Filtered NajizQA (79) QAs with scores based on answer similarity by GPT-4 as judge}
    \label{fig:NajizQA_asym.png}
\end{figure}

From the original 1358 QA pairs, we chose the filtered 79 QA + context sets. We run them with and without context and show result in Figure \ref{fig:NajizQA_asym.png}.
We notice that all models perform significantly better when the context is provided, and on closer inspection (Figure \ref{fig:najizqa-example}), we can see that the answer can easily be extracted from the context, making the context a giveaway of the answer which only measures the model's ability to recall and extract information and not legal reasoning.
% The contexts are provided, however the benchmark will only consider QA pairs without the contexts.
% We follow \verb|MMLU| and want to have questions that require memorization and recall to challenge the LLM.

%% file: latex/eval_somayah.tex
\subsection {Arabic LegalBench Evaluation}
\subsubsection{ Experiments Setup}

In total, we carried out 96 unique experiments with different prompting techniques to assess the performance of the models in legal reasoning. All prompts were in English, as this yielded better model performance. Appendix \ref{sec:LegalBench-Prompts} provides prompt examples for each technique used.

For the \textit{Contract QA} and \textit{Consumer Contract QA} datasets, we utilized the entire testing data in our experiments. However, for the larger datasets, we selected a representative sample. For each technique and dataset, we created tailored prompts. The training examples in one-shot and few-shot learning were fixed across all models for each specific dataset. We then benchmarked the performance of all models using the following four approaches: 
zero-shot learning with simple prompts where models are asked straightforward questions without extensive instructions; 
zero-shot learning with detailed instructions in the prompt;
one-shot learning; and few shot learning.

% \begin{itemize}
% \setlength{\itemsep}{-2pt}
% \item Zero-Shot Learning with Simple Prompts where models are asked straightforward questions without extensive instructions.
% \item Zero-Shot Learning with Detailed Instructions in the prompt
% \item One-Shot Learning
% \item Few Shot Learning
% \end{itemize}

\subsubsection{Results Analysis}
A comprehensive overview of model performance across all tasks and learning techniques is presented in Table \ref{tab:LegalBenchResults} in Appendix \ref{app:legalbenchR}. This table provides a summary of F1 scores for each model and task combination, offering a comparative analysis of the various approaches evaluated in this study.
 
The \textbf {consumer contract QA} task can assess the models' ability to answer questions based on a long context, in this case consumer contracts. \verb|GPT-4| with one-shot learning, and \verb|Llama3 (70B)|, with a zero-shot basic prompt, achieved the highest F1 score of 90\%. This suggests that both models can extract relevant information from consumer contracts and provide answers, even when training examples are not available or limited. 
We observed that most of the models performed well on the Contract QA task, which assessed the models capabilities to answer questions related to contracts. \verb|Command R Plus|, using few-shot learning, achieved the highest F1 score of 99\%. This high score indicates that the model can accurately understand and respond to questions about contracts when provided with a small number of training examples. However, this task proved to be the least challenging among the four tasks.
On the other hand,the privacy policy entailment task proved to be the most challenging for the LLMs across all techniques, highlighting the complexity of this task. \verb|Command R Plus|, using one-shot learning, achieved the best F1 score of 66\%. This result suggests that while the models struggle with this task, \verb|Command R Plus| is more capable of understanding of privacy policies when given a single training example.

In the final task, \textbf{Privacy Policy QA}, which evaluated the models' ability to answer questions based on privacy policies, \verb|GPT-4| with one-shot learning achieved the highest performance. This result demonstrates \verb|GPT-4|'s strong capability in extracting relevant information from privacy policies and providing accurate answers when given a single training example.

Overall, one-shot learning achieved the best results for most of the models across the various tasks. This finding suggests that providing a single example can significantly improve the models' performance in understanding and responding to questions related to legal documents such as consumer contracts and privacy policies.
% For a more detailed analysis of the experiments, please refer to Table \ref{tab:LegalBenchResults} in Appendix \ref{app:legalbenchR}

%\lipsum
\begin{figure*}[H]
    \centering
    \includegraphics[width=0.8\textwidth]{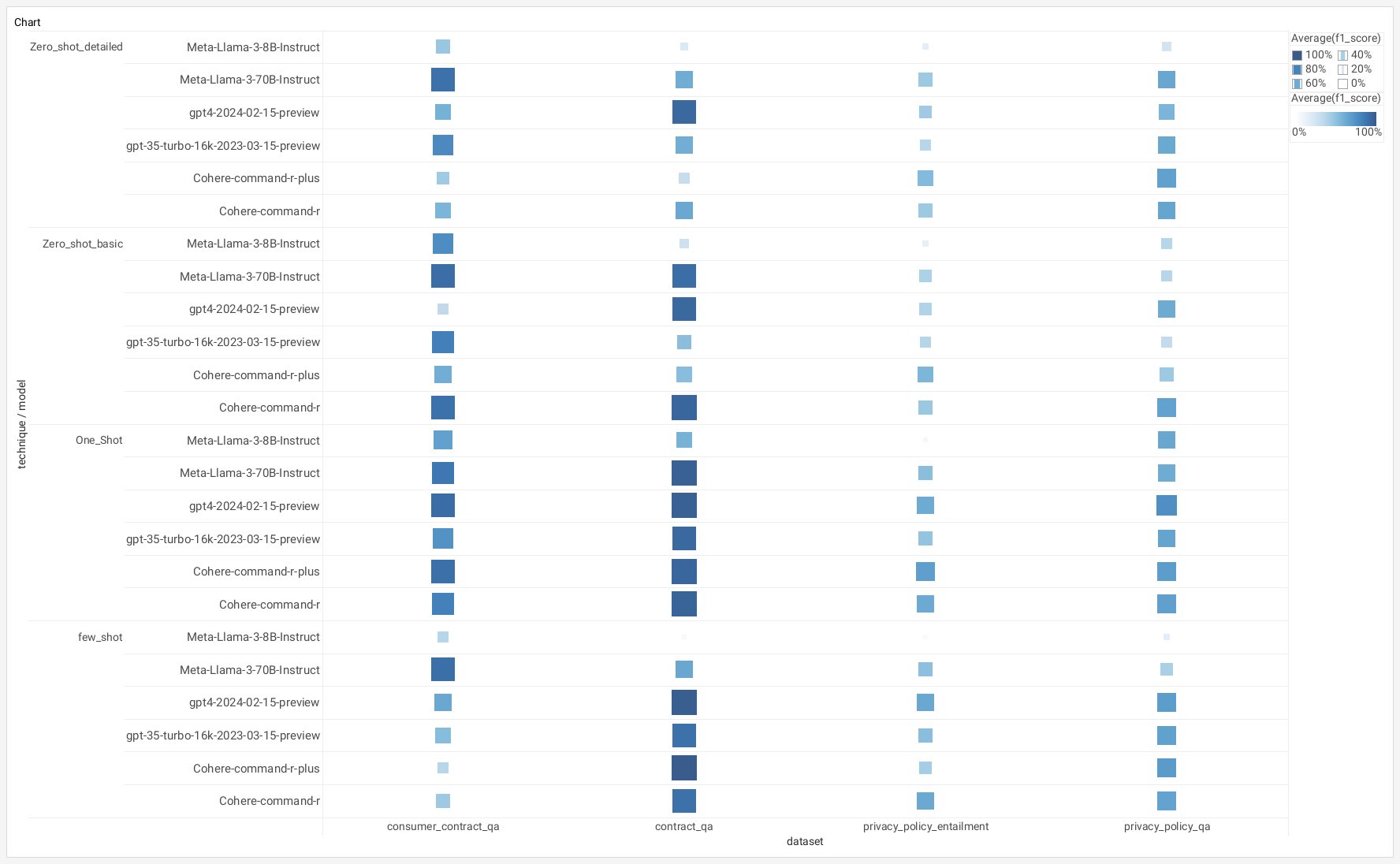}
    \caption{Experiments of Arabic LegalBench}
    \label{fig:LegalBench_Results}
\end{figure*}
%\lipsum

%% file: latex/conclusions2.tex
\section{Conclusions and Future Work}

\enlargethispage*{10pt}

We are developing an Arabic benchmark to evaluate LLMs' legal reasoning, using Saudi regulations and translated LegalBench problems. Future plans include adding more KSA regulatory documents, court cases, and judicial decisions to enhance this benchmark and promote advancements in Arabic legal AI.

% Future research will explore fine-tuning, continued pretraining, retrieval-augmented generation, and agent-based systems to navigate related legal documents, building on the foundation set by ArabLegalEval.

%% file: latex/appendix_model_info.tex
\section{LMs descriptions}
\label{sec:lm_des}

\begin{table}[H]
\caption{Models used in evaluation}
\centering
\scalebox{0.7}{

\begin{tabular}{|l |l |l|} \hline 
Model & Language & \# parameters \\ \hline 
GPT 4o & Multilingual & Undisclosed \\ \hline 
GPT 4 (0125-preview) & Multilingual & Undisclosed \\ \hline 
GPT 3.5 turbo 16k (0613) & Multilingual & Undisclosed \\ \hline 
meta-llama/Meta-Llama-3-8B-Instruct & Multilingual & 8B \\ \hline 
meta-llama/Meta-Llama-3-70B-Instruct & Multilingual & 70B \\ \hline 
%core42/jais-30b-chat-v3 & Arabic & 30B \\ \hline 
%core42/jais-13b-chat & Arabic & 13B \\ \hline 
CohereForAI/c4ai-command-r-v01 & Multilingual & 35B \\ \hline 
CohereForAI/c4ai-command-r-plus & Multilingual & 104B \\ \hline 
CohereForAI/aya-101 & Multilingual & 13B \\ \hline 
Claude 3 opus (20240229) & Multilingual & 137B \\ \hline
\end{tabular}
}
\end{table}

%% file: latex/appendix_exfigures.tex
\section{ Examples of Data Sources}
\label{sec:Examples of Data Sources}
\begin{figure}[H]
    \centering
    \includegraphics[width=0.48\textwidth]{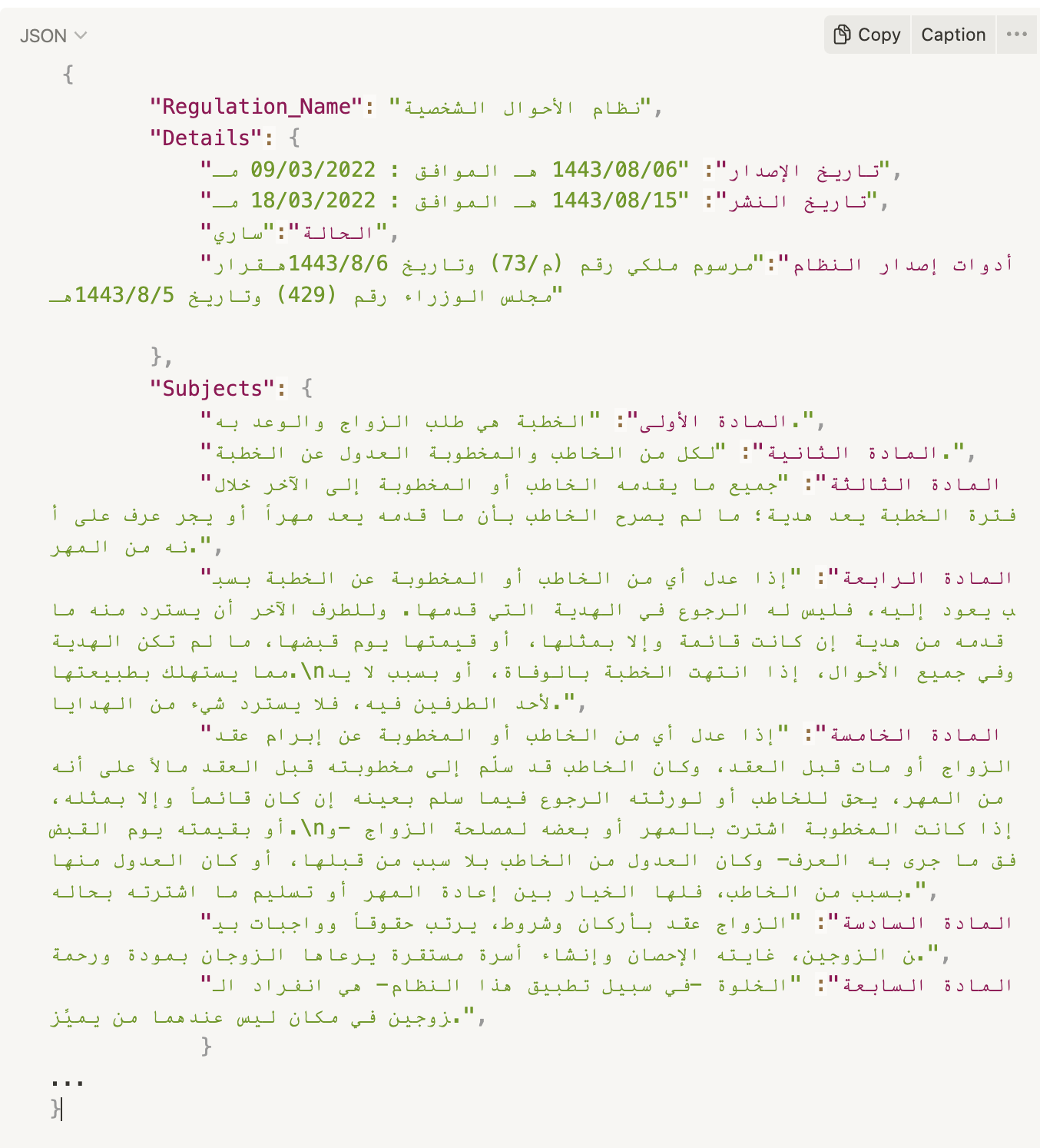}
    \caption{Example data document from Board of experts (BOE) data source}
    \label{fig:boe-example}
\end{figure}

\begin{figure}[H]
    \centering
    \includegraphics[width=0.44\textwidth]{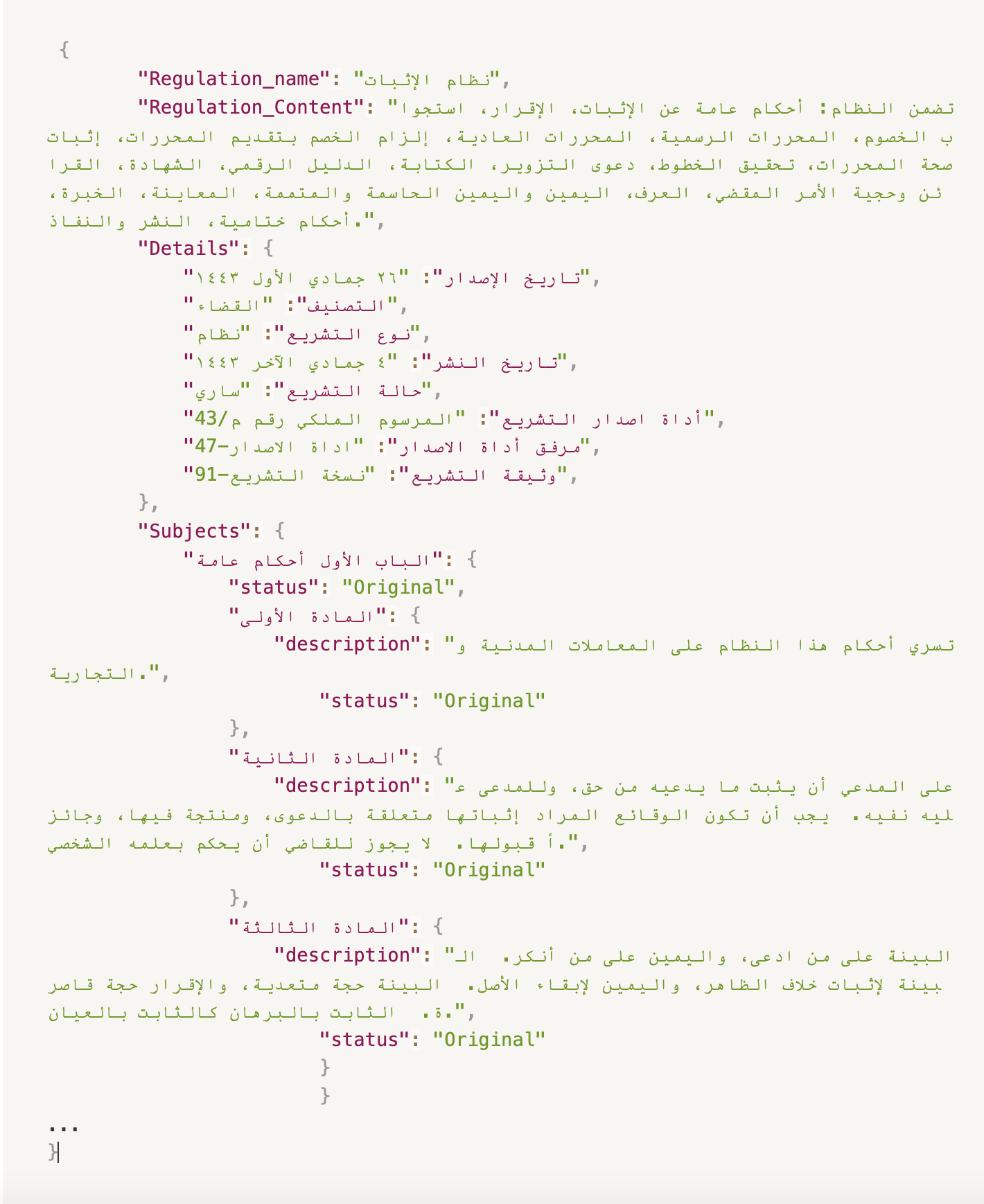}
    \caption{Example data document from Ministry~of~Justice (MOJ) data source}
    \label{fig:moj-example}
\end{figure}
\onecolumn
\begin{figure*}[h!]
    \centering
    \includegraphics[width=1\linewidth]{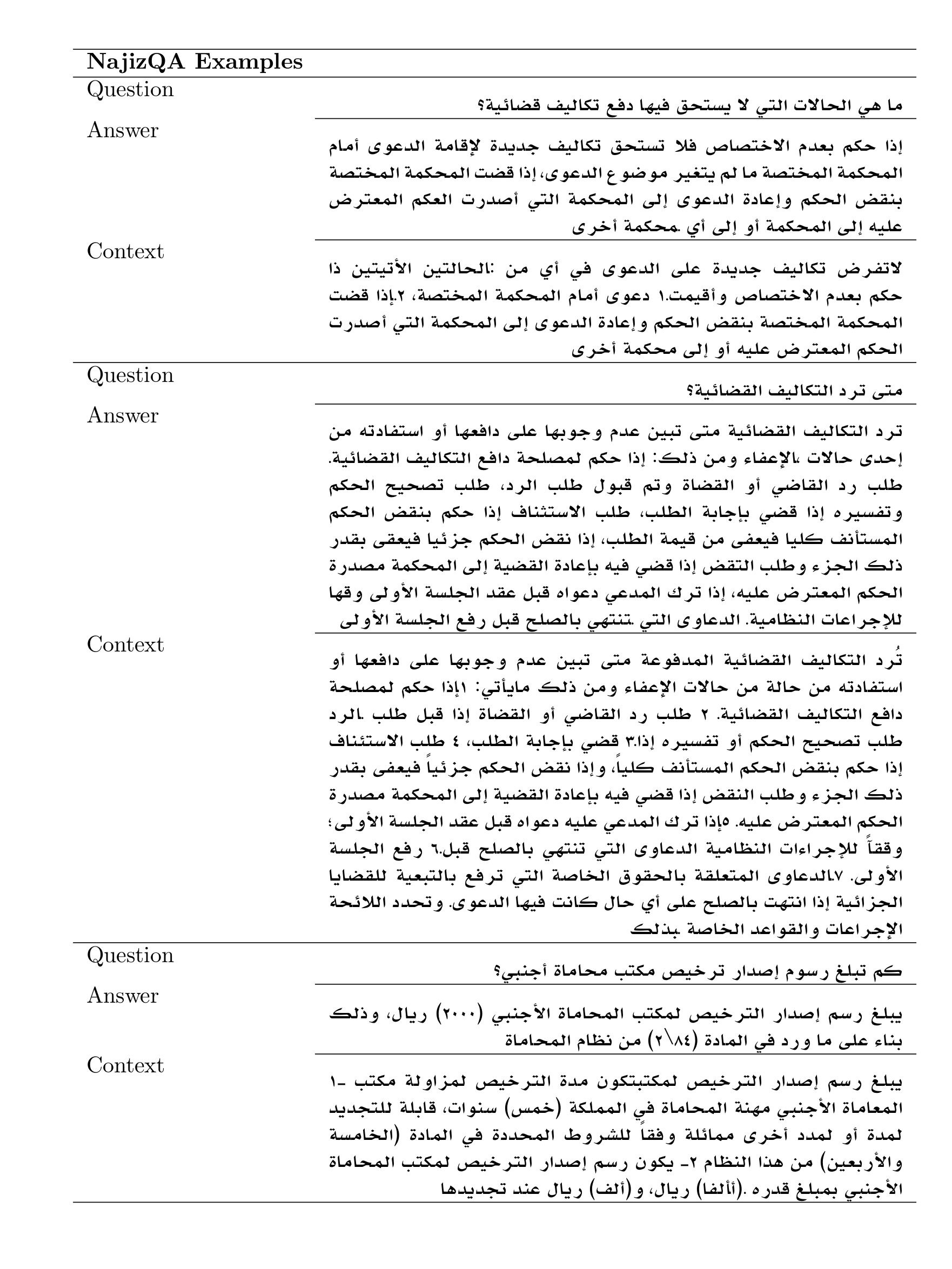}
    \caption{Examples from the NajizQA data source}
\label{fig:najizqa-example}
\end{figure*}
\twocolumn

%% file: latex/appendix_MCQs.tex
\section{MCQs Generation}

\subsection{MCQs Evaluation Criteria}\label{mcq_evaluation_criteria}
\begin{figure}[H]
    \centering
    \includegraphics[width=\linewidth]{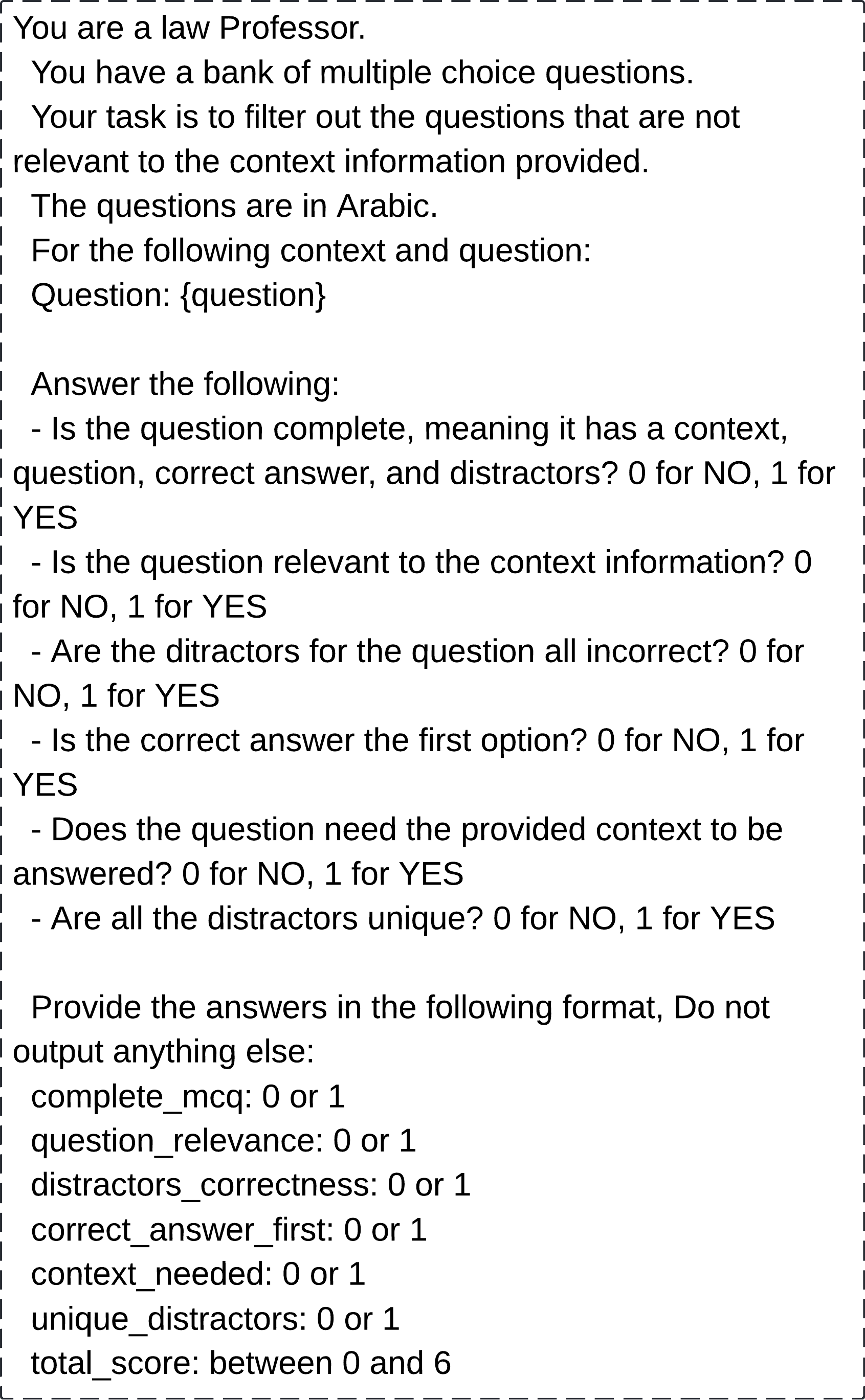}
    \caption{MCQs Evaluation Prompt}
    \label{fig:enter-label}
\end{figure}

\subsection{MCQs Example}
\begin{figure}[H]
    \centering
    \includegraphics[width=\linewidth]{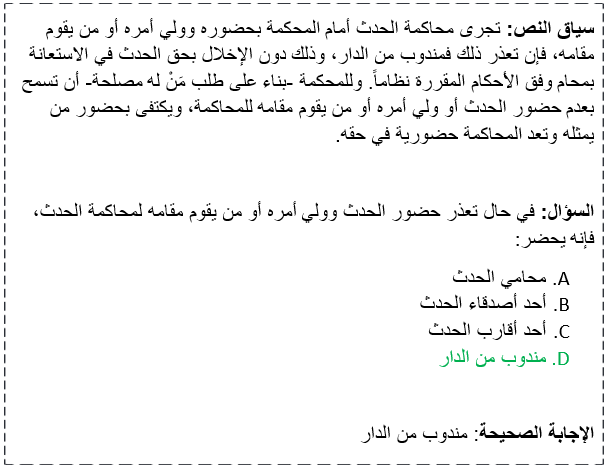}
    \caption{Example of an MCQ generated with ArabicMMLU context}
\end{figure}

\subsection{Prompts}
\begin{figure}
    \centering
    \includegraphics[width=1\linewidth]{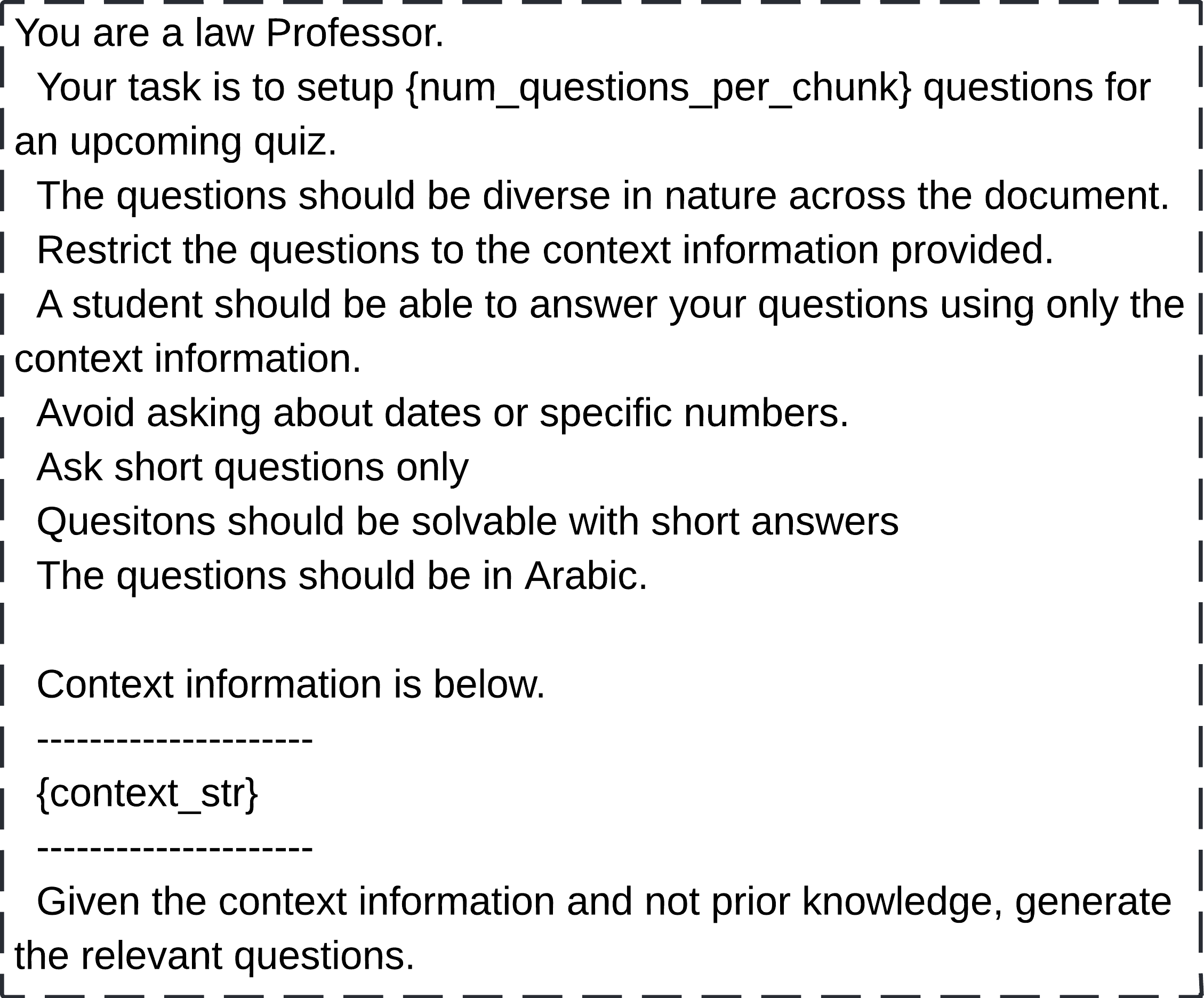}
    \caption{Question generation prompt}
    \label{fig:enter-label}
\end{figure}

\begin{figure}
    \centering
    \includegraphics[width=1\linewidth]{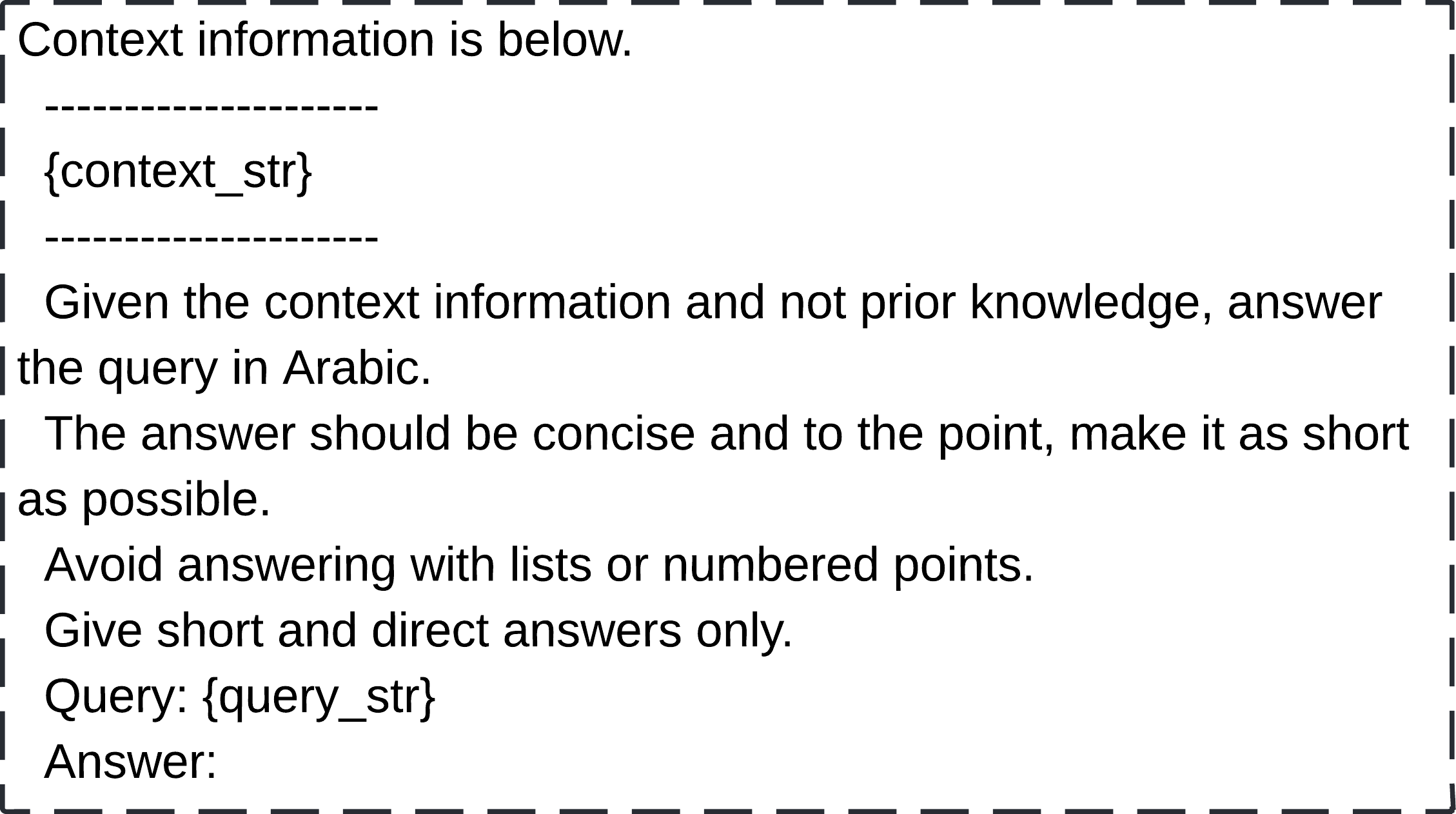}
    \caption{Question answering prompt}
    \label{fig:enter-label}
\end{figure}

\begin{figure}
    \centering
    \includegraphics[width=1\linewidth]{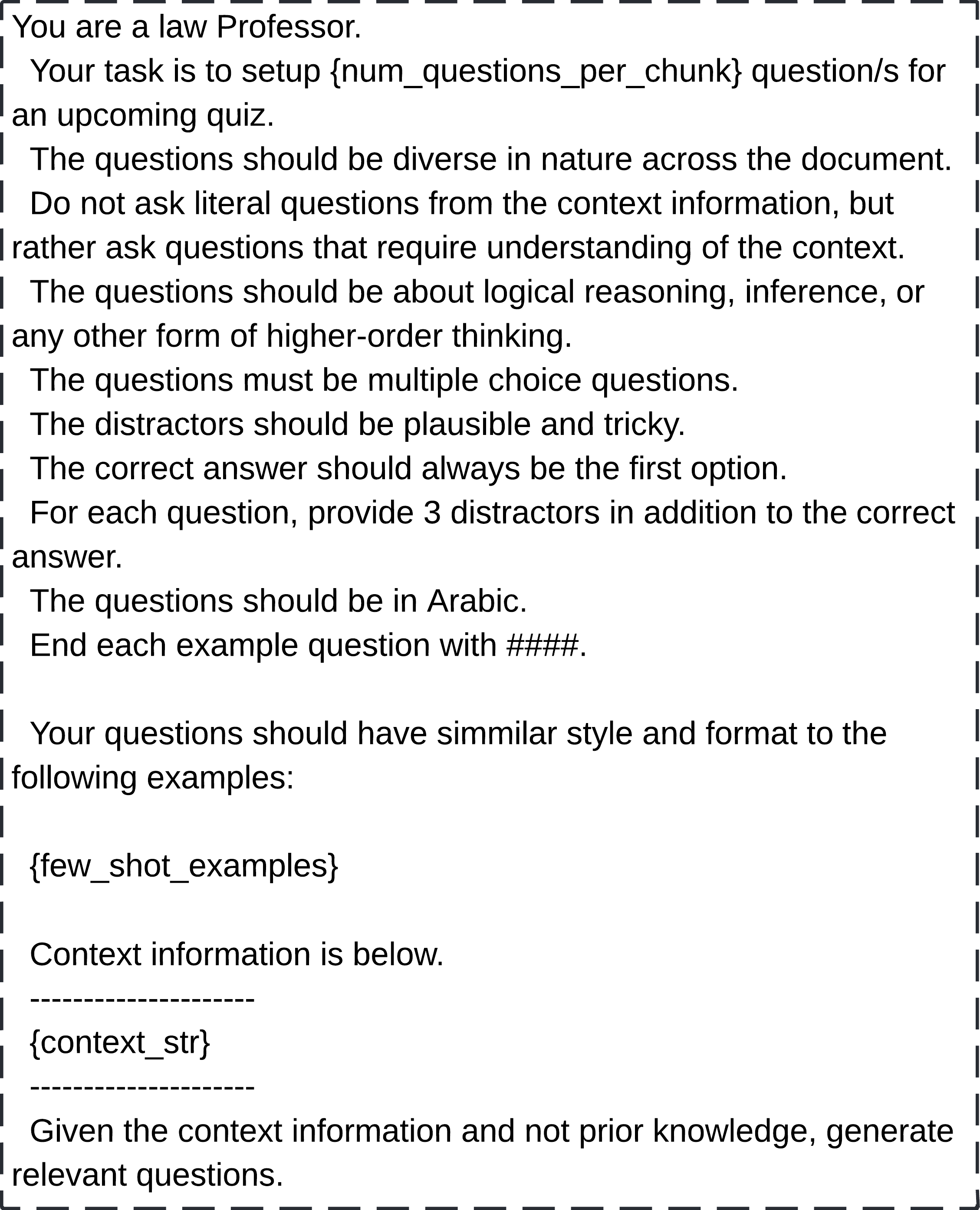}
    \caption{Prompt for generating MCQs using in-context examples}
    \label{fig:enter-label}
\end{figure}

\begin{figure}
    \centering
    \includegraphics[width=1\linewidth]{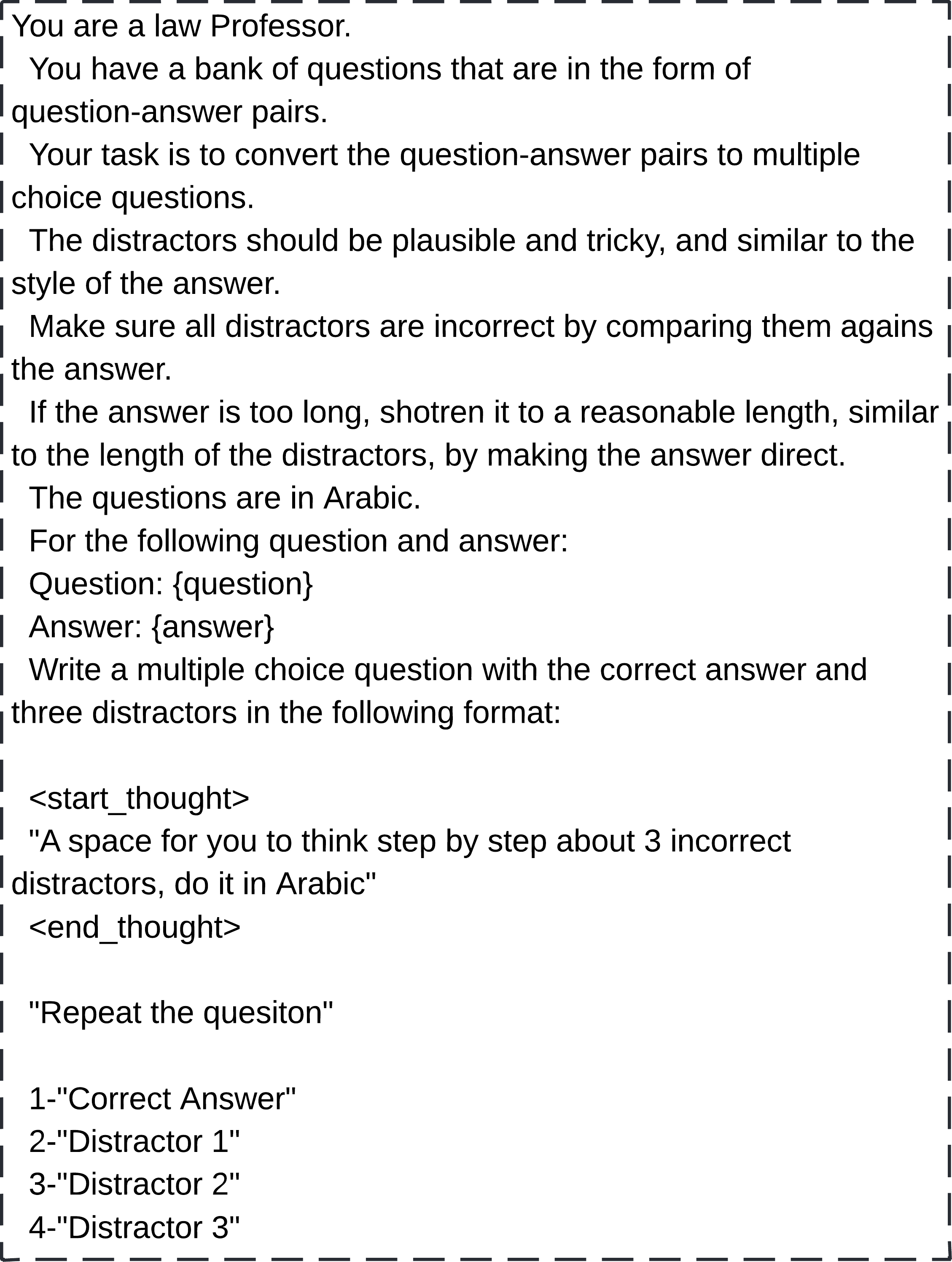}
    \caption{Converting QAs to MCQs using CoT}
    \label{fig:enter-label}
\end{figure}

%% file: latex/appendix_rouge.tex
\label{app:translation_app}
\section{Translation}
\label{sec:translation}

\subsection{ROUGE scores for UN English-Arabic translation}
\begin{table}[H]
\centering
\scalebox{0.8}{
\begin{tabular}{|c|c|c|c|}
\hline
Model & Rouge-1 & Rouge-2 & Rouge-L \\ \hline
Azure Translation Services & 0.446 & 0.242 & 0.435 \\ \hline
Google Translation API & 0.408 & 0.207 & 0.399 \\ \hline
Opus MT & \textbf{0.519} & \textbf{0.308} & \textbf{0.509} \\ \hline
GPT 4 Turbo & 0.327 & 0.138 & 0.316 \\ \hline
\end{tabular}
}
\caption{Machine Translation Results}
\label{tab:rouge_r}
\end{table}
% \subsection{GPT-4 Evaluation for the translated data}
% \vspace{-1em}
% \begin{figure}[h]
% \centering
% \includegraphics[width=0.8\linewidth]{latex/assets/GPT4_Evaluation.png}
% \caption{GPT 4 feedback on one of the Examples}
% \label{fig:GPT4_eval}
% \end{figure}

\newpage
\subsection{ GPT-4 Translation Evaluation Prompt}
\label{sec:GPT-4 Translation Evaluation}
\vspace{-1em}
\begin{figure}[H]
    \centering
    \includegraphics[width=1\columnwidth]{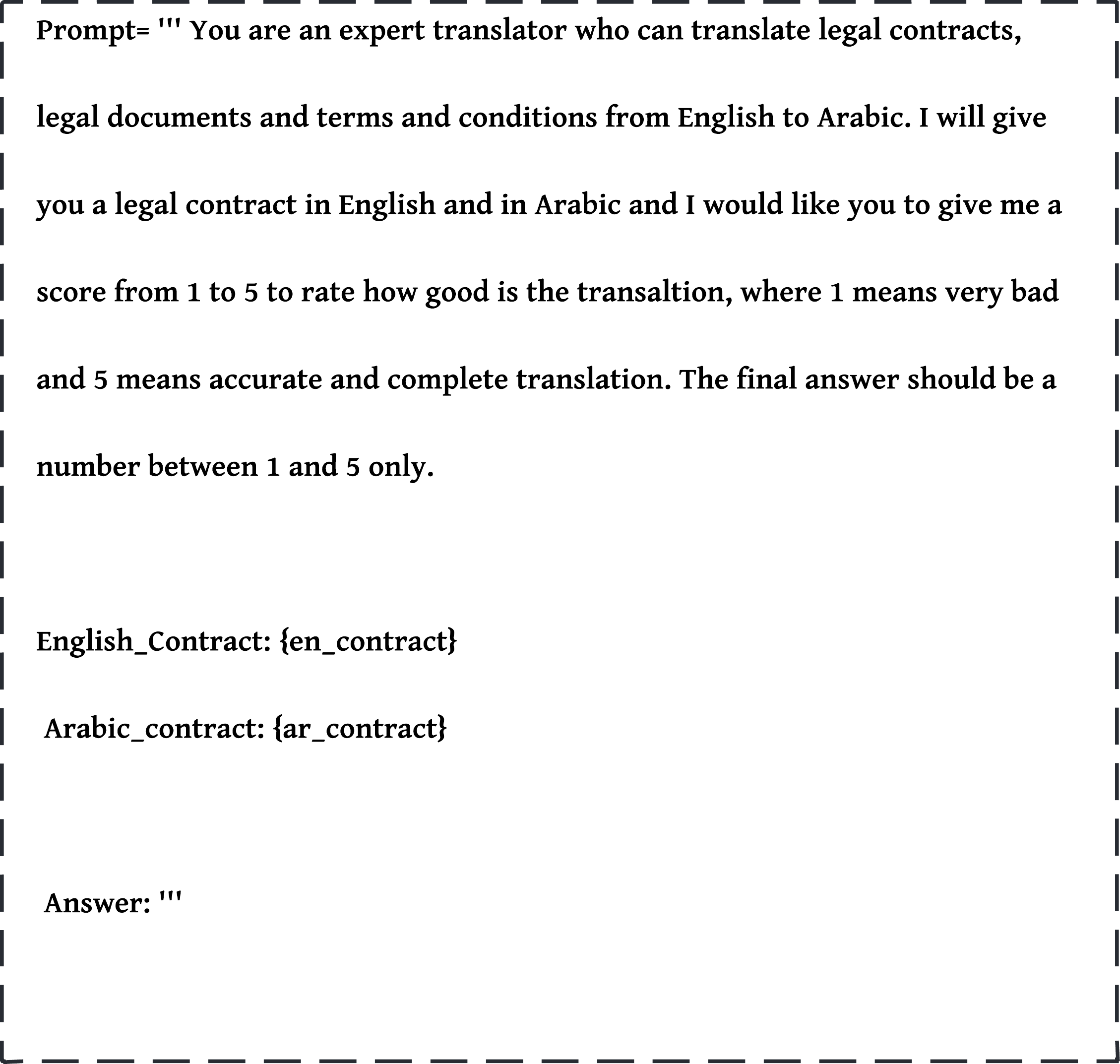}
    \caption{Prompt used with GPT-4 to evaluate the translation of LegalBench}
    \label{fig:gpt4_translationEval}
\end{figure}

\subsection{ GPT-4 Translation Result Evaluation}
\label{sec:GPT-4 Translation Evaluation}
\vspace{-1em}
\begin{figure}[H]
    \centering
    \includegraphics[width=1\columnwidth]{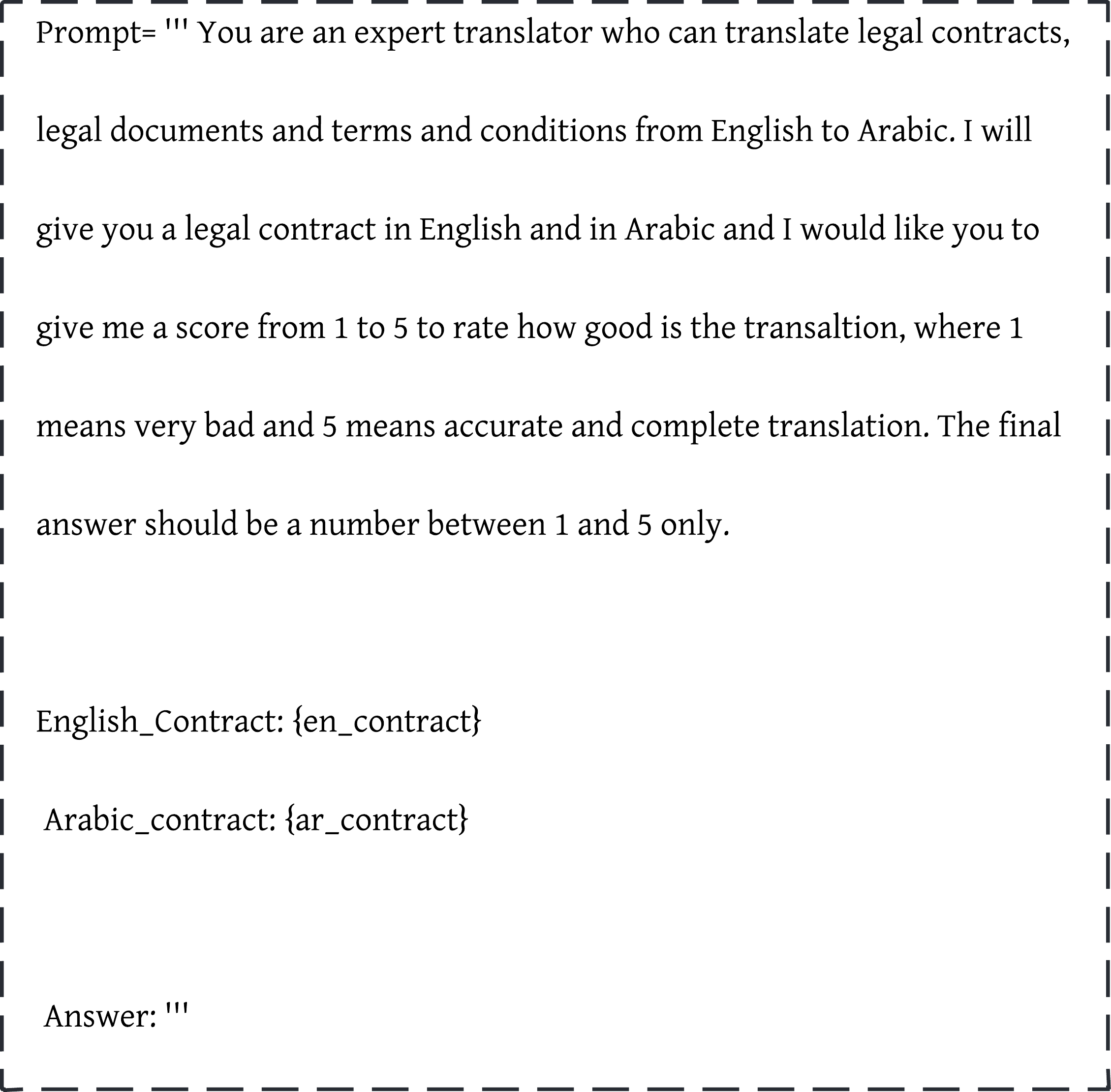}
    \caption{GPT-4 Evaluation Result}
    \label{fig:gpt4_translationResults}
\end{figure}

\subsection{Sample of the Translated Data}
    \vspace{-1em}
    \begin{figure}[h]
    \centering
    \includegraphics[width=0.5 \textwidth]{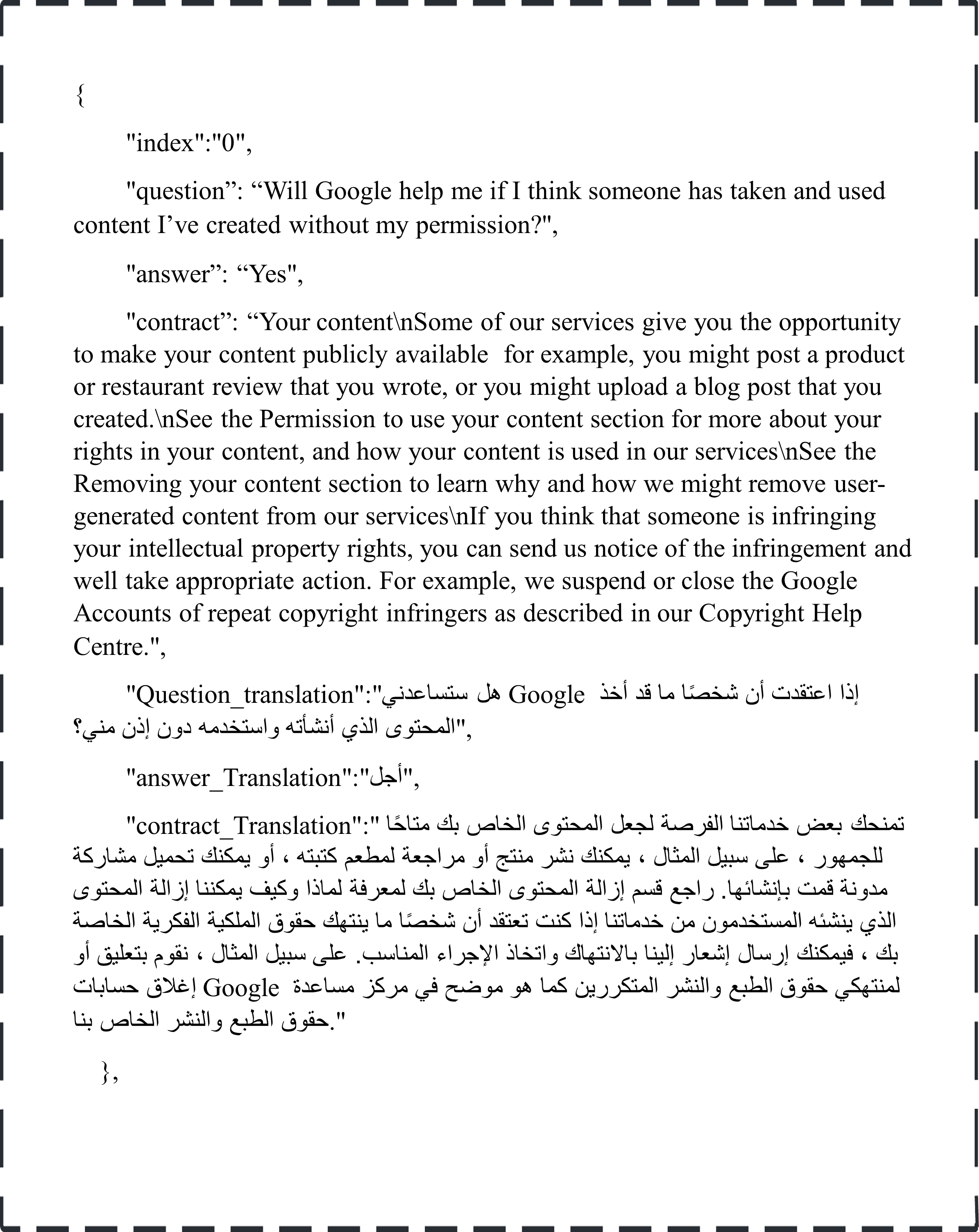}
    \caption{Sample of the Translated Data}
    \label{fig:sample_transated_Data}
\end{figure}

%% file: latex/appendix_qa.tex
\section{QA}
\begin{figure}[H]
    \centering
    \includegraphics[width=1\linewidth]{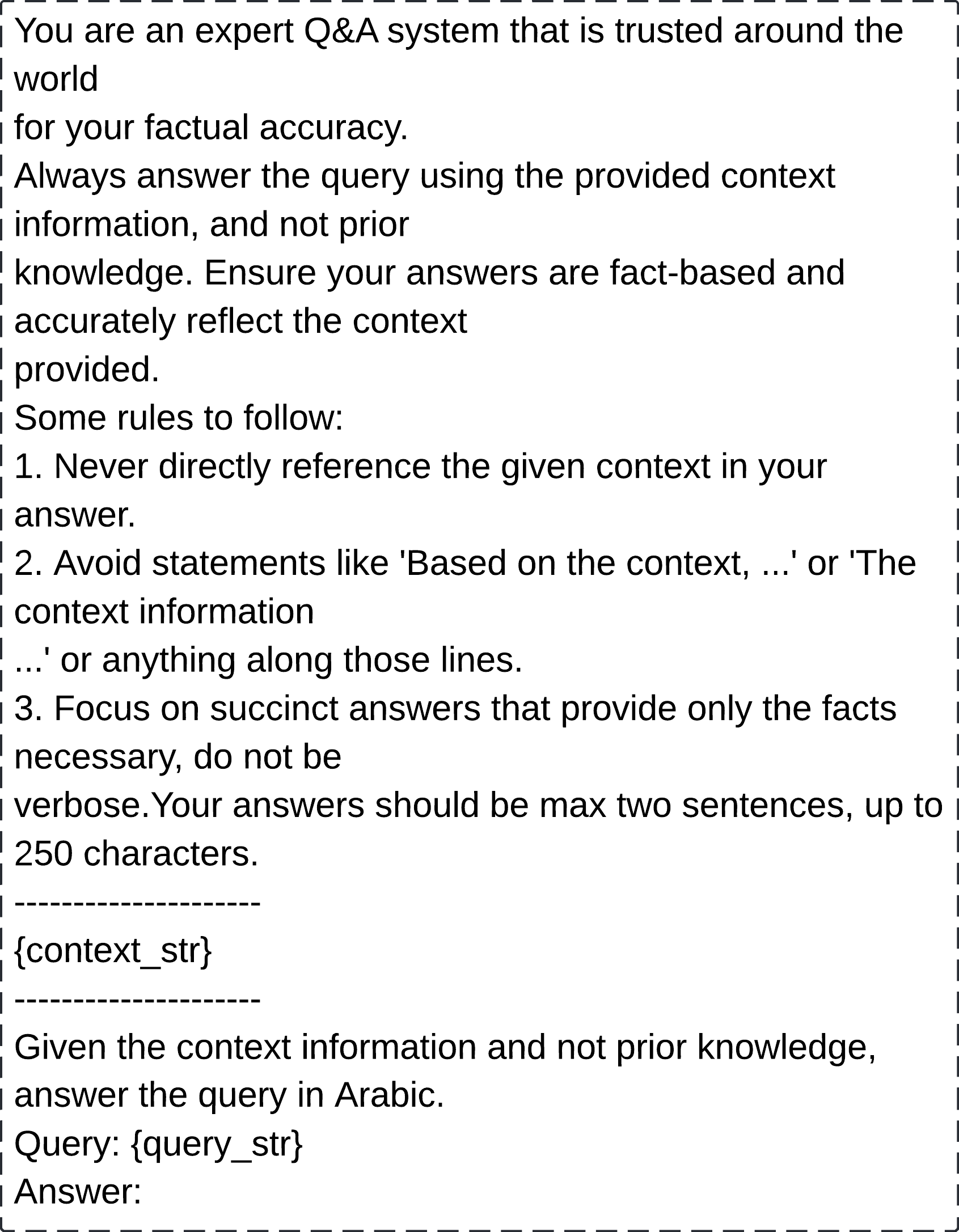}
    \caption{QA model prompt}
    \label{fig:qa-prompt}
\end{figure}
% \noindent\begin{minipage}{.45\textwidth}
% \begin{lstlisting}[caption=QA model prompt,frame=tlrb,breaklines=true]
% You are an expert Q&A system that is trusted around the world
% for your factual accuracy.

% Always answer the query using the provided context information, and not prior
% knowledge. Ensure your answers are fact-based and accurately reflect the context
% provided.

% Some rules to follow:

% 1. Never directly reference the given context in your answer.

% 2. Avoid statements like 'Based on the context, ...' or 'The context information
% ...' or anything along those lines.

% 3. Focus on succinct answers that provide only the facts necessary, do not be
% verbose.Your answers should be max two sentences, up to 250 characters.

% ---------------------

% {context_str}

% ---------------------

% Given the context information and not prior knowledge, answer the query in Arabic.

% Query: {query_str}

% Answer:

% \end{lstlisting}
% \label{fig:qa-prompt}
% \end{minipage}
\begin{figure}[H]
    \centering
    \includegraphics[width=1\linewidth]{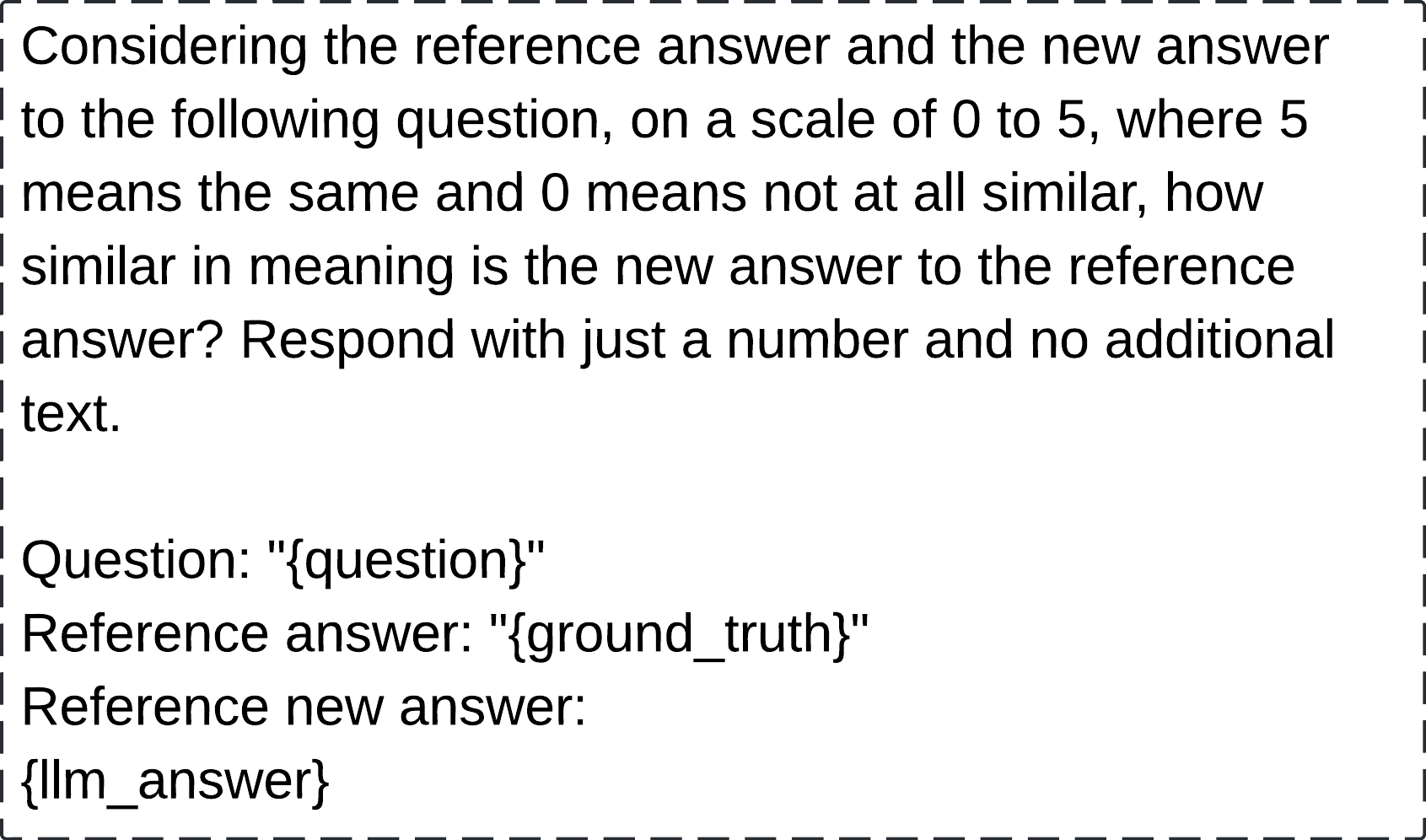}
    \caption{LM Judge answer similarity prompt}
    \label{fig:llm-judge-prompt}
\end{figure}

% \noindent\begin{minipage}{.45\textwidth}
% \begin{lstlisting}[caption=LLM Judge answer similarity prompt,frame=tlrb,breaklines=true]
% Considering the reference answer and the new answer
% to the following question, on a scale of 0 to 5, where 5 means the same and 0
% means not at all similar, how similar in meaning is the new answer to the reference
% answer? Respond with just a number and no additional text.

% Question: "{question}"

% Reference answer: "{ground_truth}"

% Reference new answer:
% {llm_answer}
% \end{lstlisting}
% \label{fig:llm-judge-prompt}
% \end{minipage}

%% file: latex/appendix_arbmmlu.tex
\onecolumn
\section{MCQs Evaluation}
\label{sec:arabicmmlu_opt}
\begin{minipage}{0.5\columnwidth}
\subsection{DSPy Signatures}
\begin{figure}[H]
    \centering
    \includegraphics[width=0.95\columnwidth]{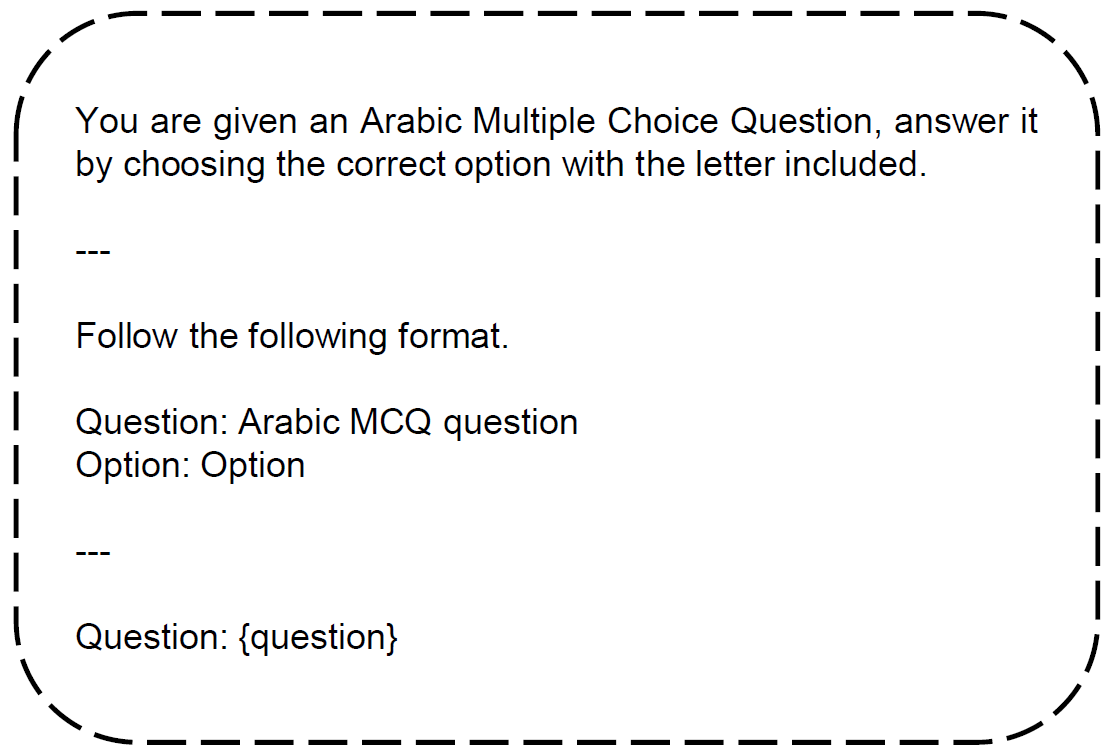}
    \caption{DSPy signature for few-shot prompt}
    \label{fig:dspyfew.png}
\end{figure}
\begin{figure}[H]
    \centering
    \includegraphics[width=0.95\columnwidth]{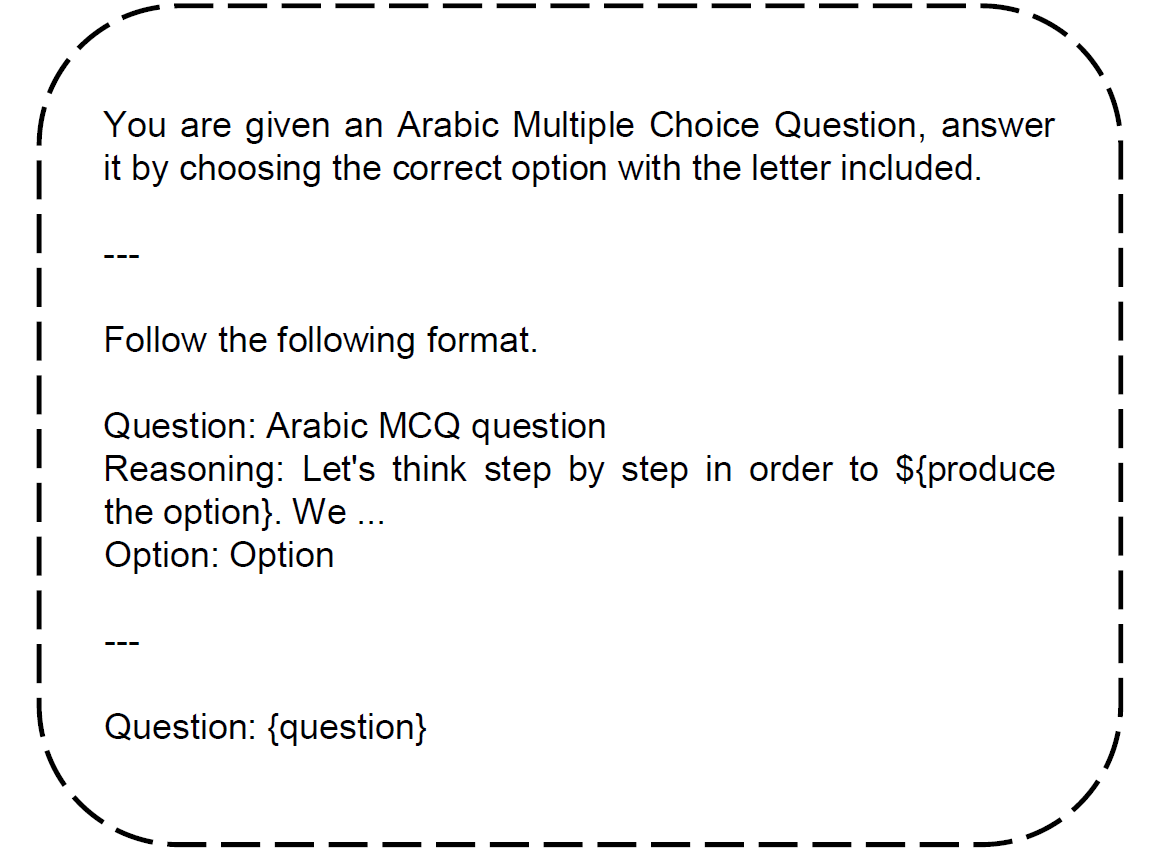}
    \caption{DSPy signature for few-shot with CoT prompt}
    \label{fig:dspycot.png}
\end{figure}
\subsection{ArabicMMLU Results with DSPy optimization}
Table \ref{tab:armmlu} displays the performance of models on ArabicMMLU's law and political science subsets with optimized prompts. The prompts were optimized with 15 examples as labeled inputs from law and political science subsets, respectively. Therefore, the accuracies were computed using the remaining 299 and 195 samples from the respective law and political science sets. 
\end{minipage}

\begin{table*}[h]
\begin{center}
\caption{Experimental results on Law and Political Science subsets of ArabicMMLU}
\label{tab:armmlu}
\scalebox{0.75}{
\begin{tabular}{|c|c|c|c|c|c|c|}
\hline
\multicolumn{1}{|l|}{}      & \multicolumn{1}{l|}{} & \multicolumn{1}{l|}{Original} & \multicolumn{1}{l|}{Few-shot} & \multicolumn{1}{l|}{\shortstack{Few-shot \\ (GPT-4 Teacher)}} & \multicolumn{1}{l|}{CoT Few-shot} & \multicolumn{1}{l|}{\shortstack{CoT Few-shot \\ (GPT-4 Teacher)}} \\ \hline
\multirow{2}{*}{GPT-4}      & Law                   & 73.60\%                       & 74.90\%                       & -                                          & \textbf{77.30\%}                  & -                                              \\ \cline{2-7} 
                            & Political Science     & 70.80\%                       & \textbf{75.90\%}              & -                                          & 70.80\%                           & -                                              \\ \hline
\multirow{2}{*}{GPT-4o}     & Law                   & 68.20\%                       & \textbf{82.90\%}              & -                                          & 81.90\%                           & -                                              \\ \cline{2-7} 
                            & Political Science     & 71.30\%                       & \textbf{74.90\%}              & -                                          & 73.80\%                           & -                                              \\ \hline
\multirow{2}{*}{Command R}  & Law                   & 56.90\%                       & 61.50\%                       & 65.20\%                                       & 69.20\%                           & \textbf{69.90\%}                                  \\ \cline{2-7} 
                            & Political Science     & 65.10\%                       & 67.20\%                       & 67.70\%                                       & \textbf{72.80\%}                  & 66.70\%                                           \\ \hline
\multirow{2}{*}{Command R+} & Law                   & 56.50\%                       & 61.20\%                       & 65.60\%                                       & \textbf{72.20\%}                  & 68.20\%                                           \\ \cline{2-7} 
                            & Political Science     & 65.10\%                       & 66.70\%                       & 67.70\%                                       & \textbf{72.30\%}                  & 66.20\%                                           \\ \hline
\multirow{2}{*}{Llama3 8B}  & Law                   & 56.90\%                       & 61.50\%                       & 65.90\%                                       & \textbf{73.20\%}                  & 69.20\%                                           \\ \cline{2-7} 
                            & Political Science     & 65.10\%                       & 67.20\%                       & 67.20\%                                       & \textbf{73.30\%}                  & 65.10\%                                           \\ \hline
\multirow{2}{*}{Llama3 70B} & Law                   & 56.60\%                       & 61.50\%                       & 65.90\%                                       & \textbf{73.90\%}                  & 67.60\%                                           \\ \cline{2-7} 
                            & Political Science     & 65.10\%                       & 67.20\%                       & 67.70\%                                       & \textbf{70.30\%}                  & 66.70\%                                           \\ \hline
\multirow{2}{*}{Aya101}     & Law                   & 45.80\%                       & 42.80\%                       & 19.10\%                                       & 21.70\%                           & \textbf{47.10\%}                                  \\ \cline{2-7} 
                            & Political Science     & 47.20\%                       & 50.80\%                       & 51.30\%                                       & \textbf{51.80\%}                  & 26.10\%                                           \\ \hline
\end{tabular}
}
\end{center}
\end{table*}

%% file: latex/appendix_ArabicLB_Results.tex
\clearpage
\begin{minipage}{0.5\columnwidth}
\section{Details for Arabic LegalBench Experiments Results}
\label{app:legalbenchR}
\end{minipage}
\begin{table*}[h]
\caption{Experiments Results on Arabic LegalBench Data}
\label{tab:LegalBenchResults}
\scalebox{0.65}{
\begin{tabular}{|c|c|c|c|c|c|} \hline
\textbf{Technique}                    & \textbf{Model}                      & \textbf{consumer\_contract\_qa} & \textbf{contract\_qa} & \textbf{privacy\_policy\_entailment} & \textbf{privacy\_policy\_qa} \\ \hline
\multirow{6}{*}{few\_shot}            & Cohere-command-r                    & 43\%                            & 89\%                  & 61\%                                 & 64\%                         \\ \cline{2-6} 
                                      & Cohere-command-r-plus               & 35\%                            & \textbf{99\%}         & 41\%                                 & \textbf{68\%}                \\ \cline{2-6} 
                                      & gpt-35-turbo-16k-2023-03-15-preview & 51\%                            & 89\%                  & 50\%                                 & 66\%                         \\ \cline{2-6} 
                                      & gpt4-2024-02-15-preview             & 62\%                            & 97\%                  & \textbf{62\%}                        & 67\%                         \\ \cline{2-6} 
                                      & Meta-Llama-3-70B-Instruct           & \textbf{89\%}                   & 62\%                  & 49\%                                 & 40\%                         \\ \cline{2-6} 
                                      & Meta-Llama-3-8B-Instruct            & 35\%                            & 1\%                   & 0\%                                  & 14\%                         \\ \hline
\multirow{6}{*}{One\_Shot}            & Cohere-command-r                    & 81\%                            & 95\%                  & 61\%                                 & 66\%                         \\ \cline{2-6} 
                                      & Cohere-command-r-plus               & 89\%                            & 94\%                  & \textbf{66\%}                        & 68\%                         \\ \cline{2-6} 
                                      & gpt-35-turbo-16k-2023-03-15-preview & 73\%                            & 92\%                  & 47\%                                 & 64\%                         \\ \cline{2-6} 
                                      & gpt4-2024-02-15-preview             & \textbf{90\%}                   & \textbf{96\%}         & 60\%                                 & \textbf{74\%}                \\ \cline{2-6} 
                                      & Meta-Llama-3-70B-Instruct           & 86\%                            & \textbf{96\%}         & 50\%                                 & 60\%                         \\ \cline{2-6} 
                                      & Meta-Llama-3-8B-Instruct            & 65\%                            & 56\%                  & 4\%                                  & 62\%                         \\ \hline
\multirow{6}{*}{Zero\_shot\_basic}    & Cohere-command-r                    & 88\%                            & \textbf{94\%}         & 44\%                                 & \textbf{65\%}                \\ \cline{2-6} 
                                      & Cohere-command-r-plus               & 59\%                            & 50\%                  & \textbf{54\%}                        & 43\%                         \\ \cline{2-6} 
                                      & gpt-35-turbo-16k-2023-03-15-preview & 82\%                            & 50\%                  & 36\%                                 & 30\%                         \\ \cline{2-6} 
                                      & gpt4-2024-02-15-preview             & 31\%                            & 92\%                  & 37\%                                 & 60\%                         \\ \cline{2-6} 
                                      & Meta-Llama-3-70B-Instruct           & \textbf{90\%}                   & 90\%                  & 38\%                                 & 35\%                         \\ \cline{2-6} 
                                      & Meta-Llama-3-8B-Instruct            & 75\%                            & 26\%                  & 12\%                                 & 35\%                         \\ \hline
\multirow{6}{*}{Zero\_shot\_detailed} & Cohere-command-r                    & 55\%                            & 62\%                  & 44\%                                 & 63\%                         \\ \cline{2-6} 
                                      & Cohere-command-r-plus               & 43\%                            & 29\%                  & \textbf{52\%}                        & \textbf{65\%}                \\ \cline{2-6} 
                                      & gpt-35-turbo-16k-2023-03-15-preview & 77\%                            & 59\%                  & 35\%                                 & 62\%                         \\ \cline{2-6} 
                                      & gpt4-2024-02-15-preview             & 57\%                            & \textbf{92\%}         & 43\%                                 & 55\%                         \\ \cline{2-6} 
                                      & Meta-Llama-3-70B-Instruct           & \textbf{89\%}                   & 59\%                  & 44\%                                 & 62\%                         \\ \cline{2-6} 
                                      & Meta-Llama-3-8B-Instruct            & 45\%                            & 18\%                  & 13\%                                 & 23\%                         \\ \hline
\end{tabular}
}
\end{table*}
\twocolumn

%% file: latex/appendix_LegalBench_Prompts.tex
\label{app:legalbench-prompts}
\section{ Araic LegalBench Prompts}
\label{sec:LegalBench-Prompts}

\begin{figure}[H]
    \centering
    \includegraphics[width=\columnwidth]{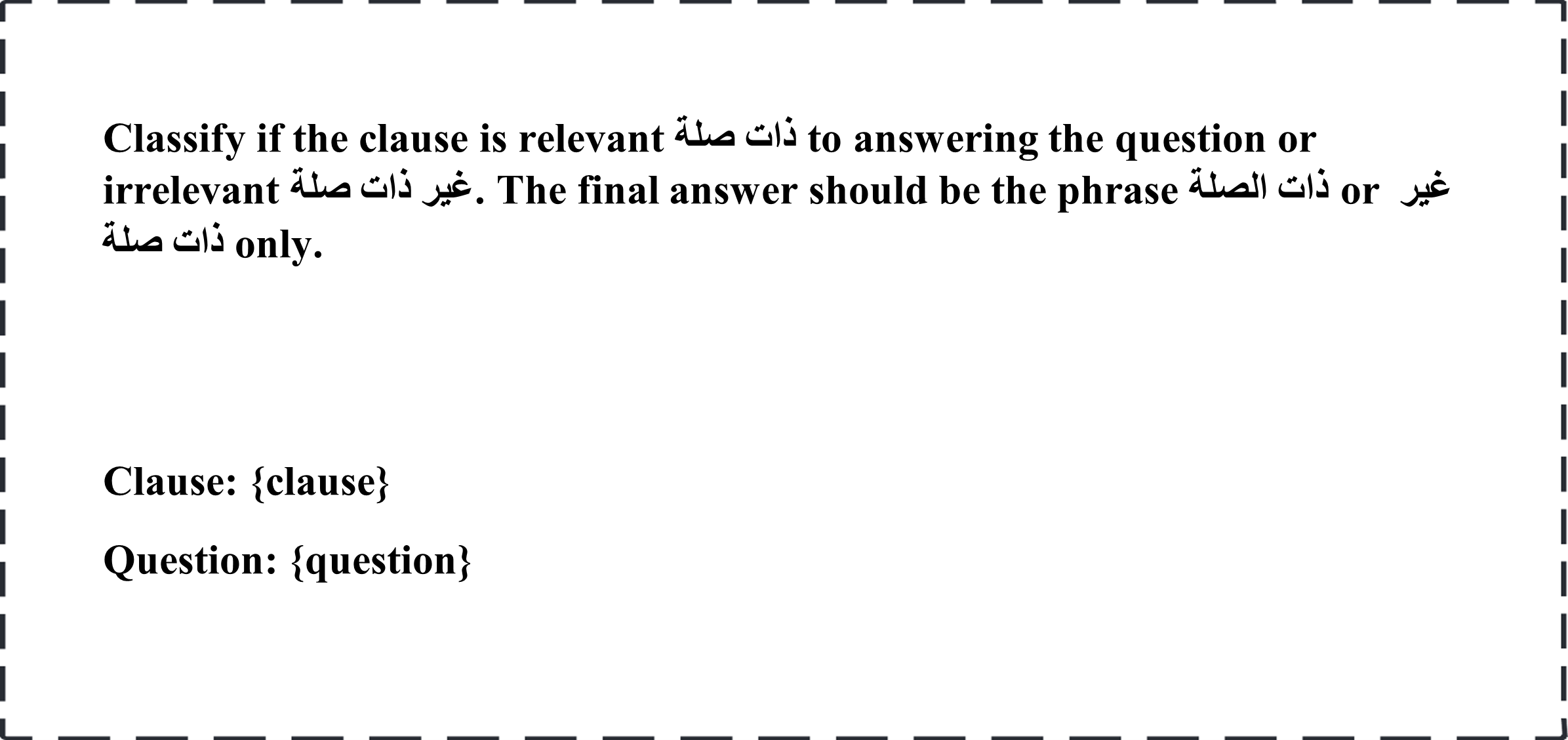}
    \caption{Zero-Shot Basic Technique Example}
    \label{fig:Zero-shot}
\end{figure}

\begin{figure}[H]
    \centering
    \includegraphics[width=\columnwidth]{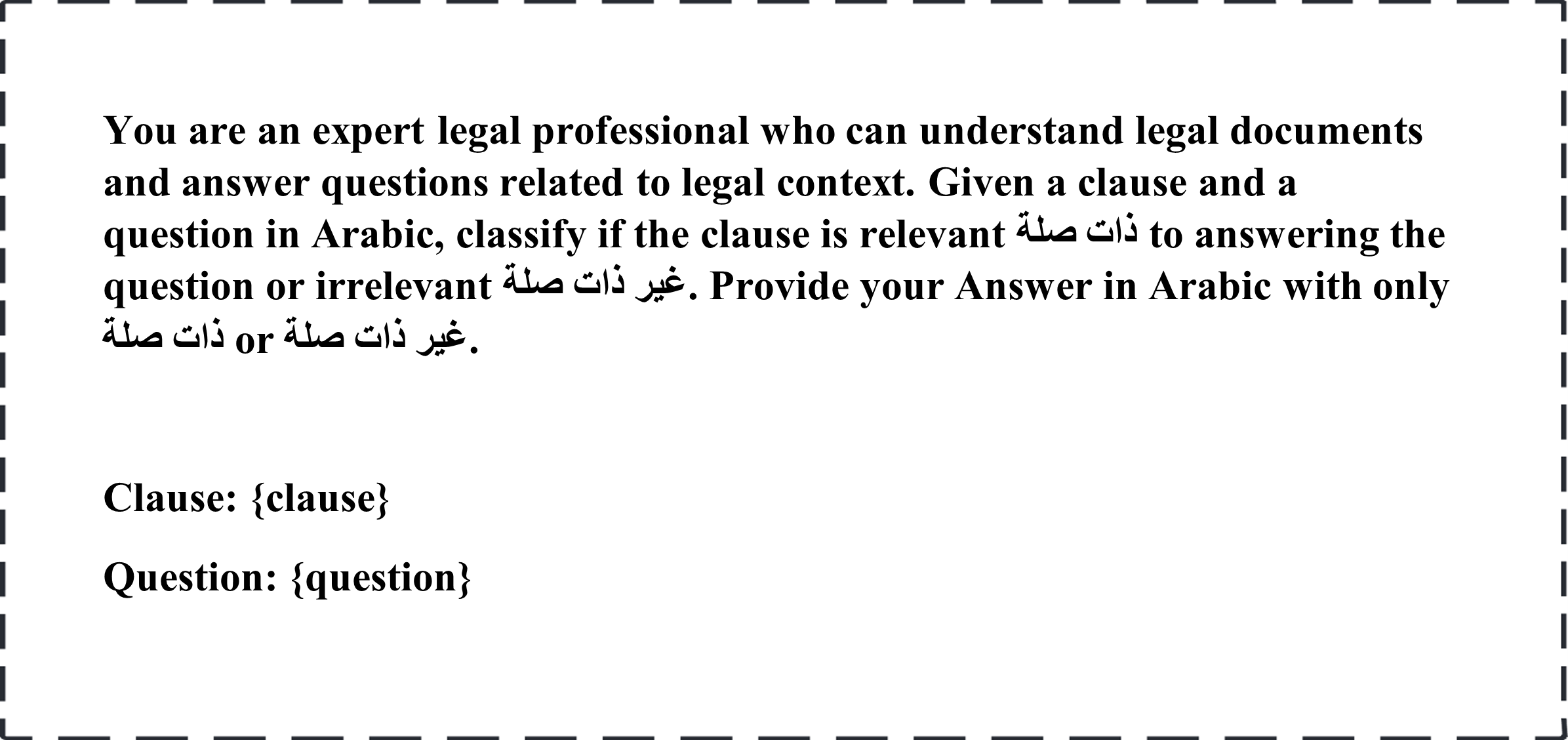}
    \caption{Zero-Shot Detailed Technique Example}
    \label{fig:Zero-shot-detailed}
\end{figure}
\begin{figure}[H]
    \centering
    \includegraphics[width=\columnwidth]{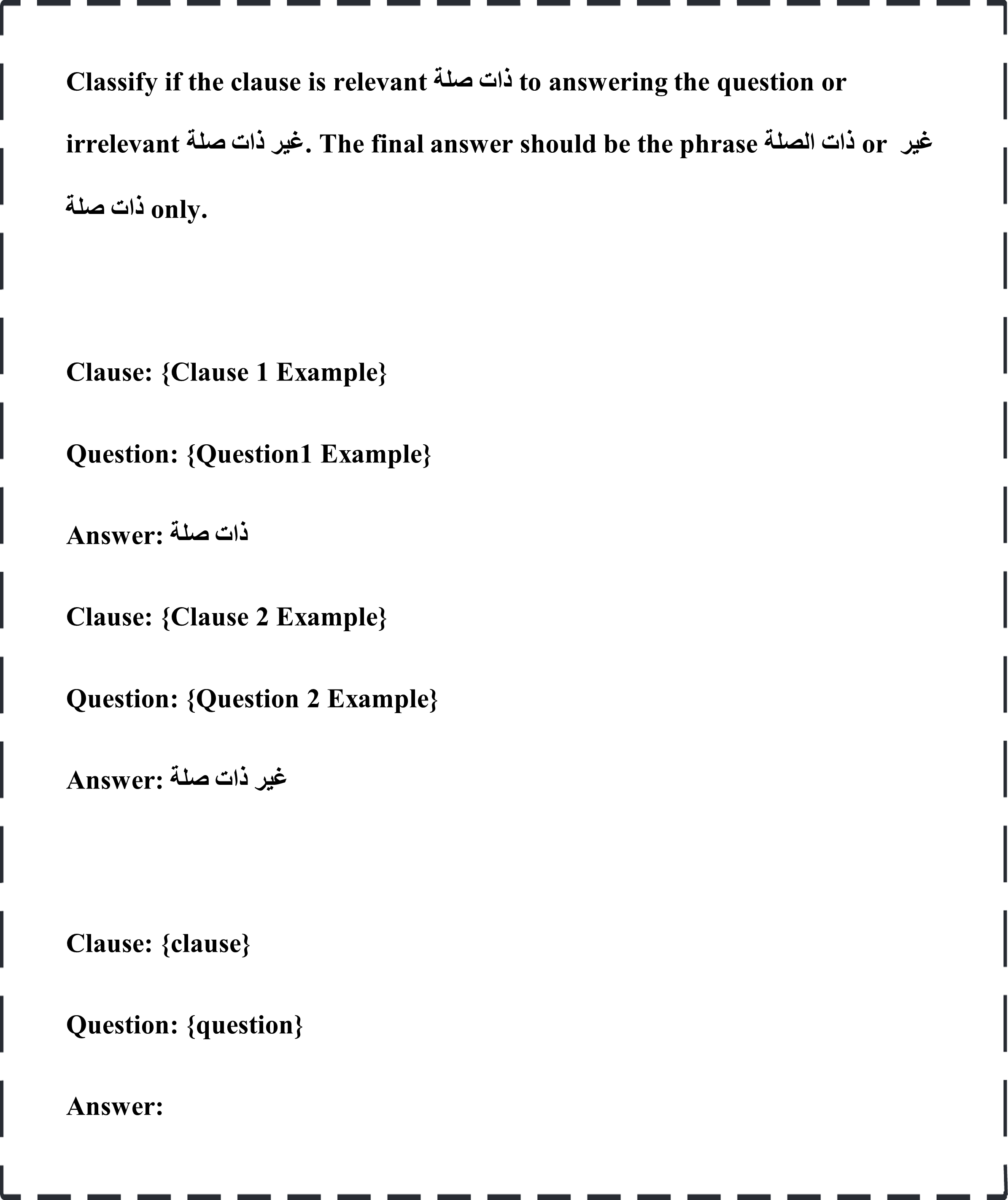}
    \caption{One-Shot Basic Technique Example}
    \label{fig:One-shot}
\end{figure}
\begin{figure}[H]
    \centering
    \includegraphics[width=\columnwidth]{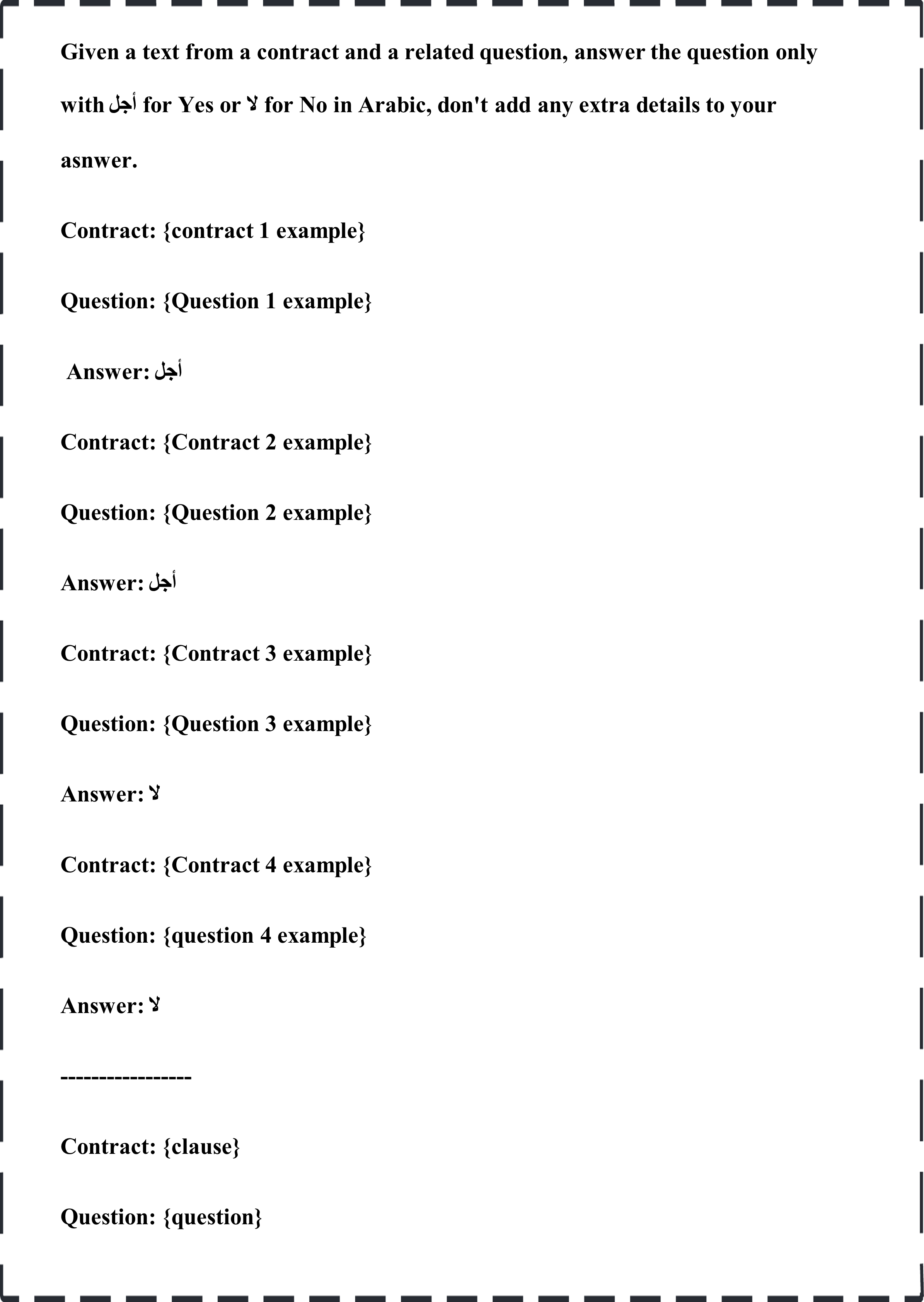}
    \caption{Zero-Shot Basic Technique Example}
    \label{fig:Few-shots}
\end{figure}

%% file: latex/appendix_quantAnalysis.tex
\newpage
\section{Dataset Quantitative Analysis}
\label{sec:quant-analysis}

\subsection{MCQs}

\begin{figure}[H]
    \centering
    \includegraphics[width=1\linewidth]{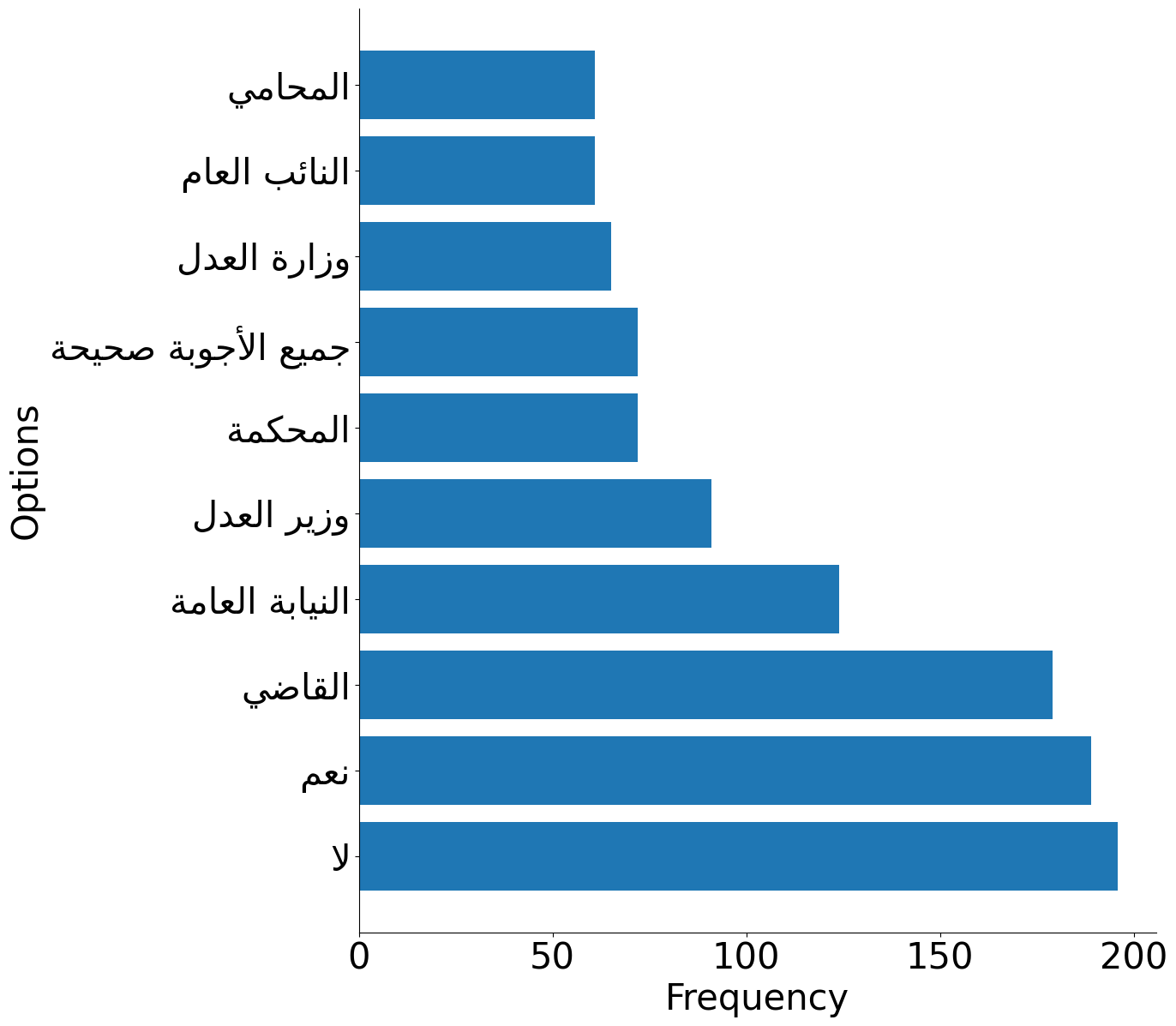}
    \caption{Top 10 most frequent choices}
    \label{fig:enter-label}
\end{figure}

\begin{figure}[H]
    \centering
    \includegraphics[width=1\linewidth]{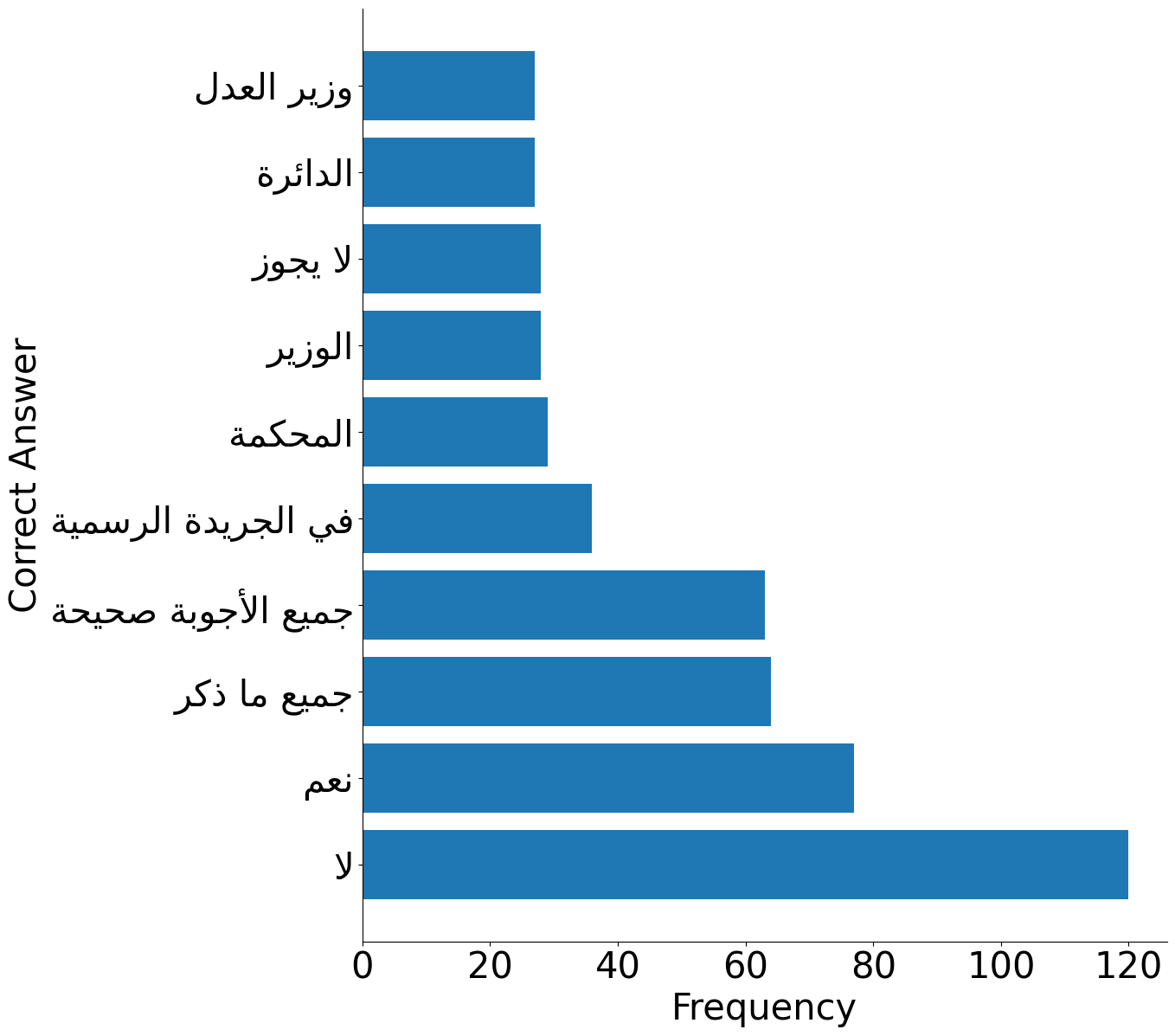}
    \caption{Top 10 most frequent correct answers}
    \label{fig:enter-label}
\end{figure}

\begin{table}[H]
    \centering
    \begin{tabular}{lccc}
        \hline
        & Min Length & Max Length & Avg Length \\
        \hline
        Context & 4 & 1497 & 46.0 \\
        Question & 2 & 48 & 12.0 \\
        \hline
    \end{tabular}
    \caption{Length of context documents and questions (words)}
    \label{table:stats}
\end{table}

\begin{figure}[H]
    \centering
    \includegraphics[width=1\linewidth]{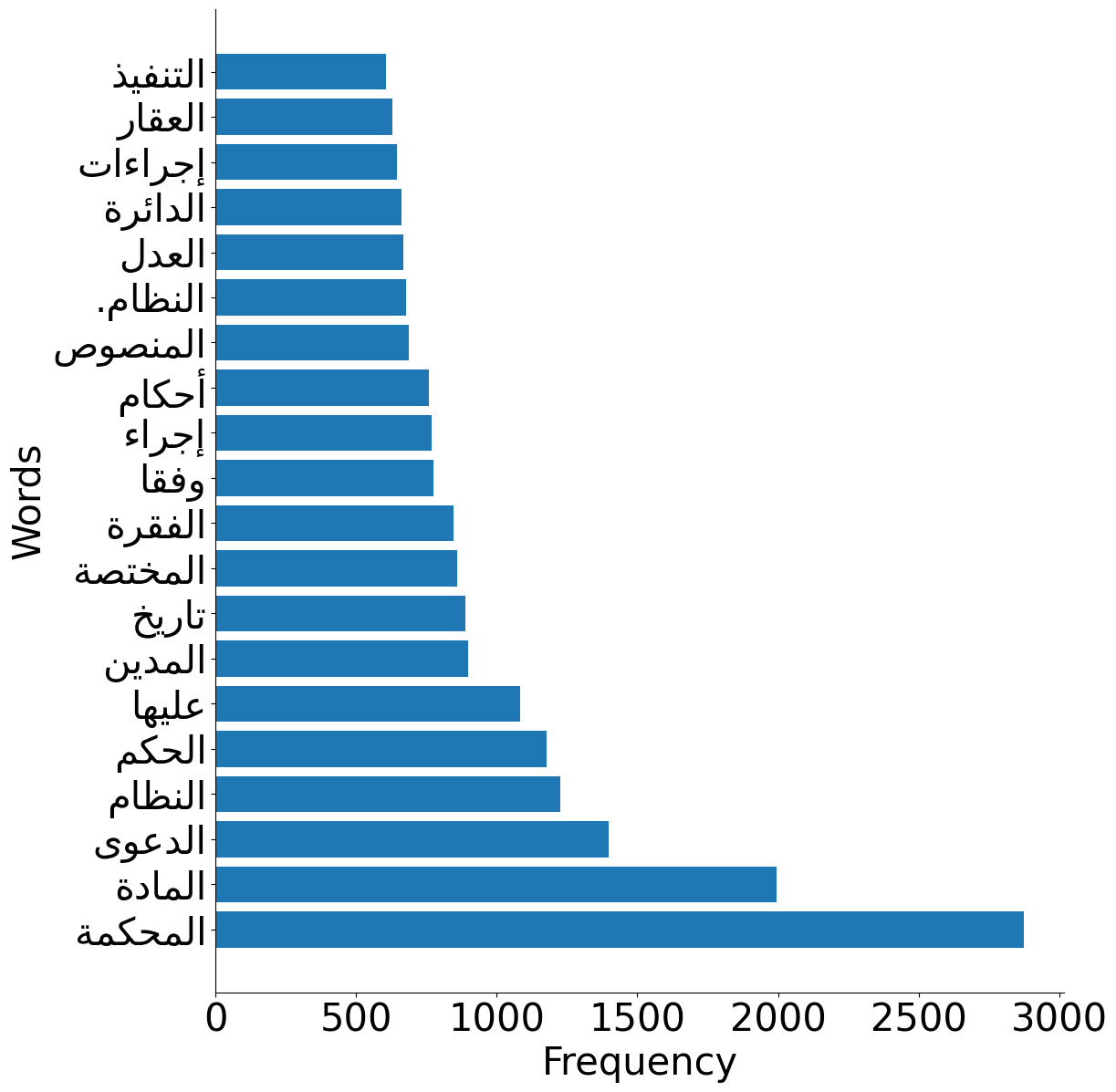}
    \caption{Top 20 frequency of words in the context of the MCQs}
    \label{fig:enter-label}
\end{figure}
\begin{figure}[H]
    \centering
    \includegraphics[width=1\linewidth]{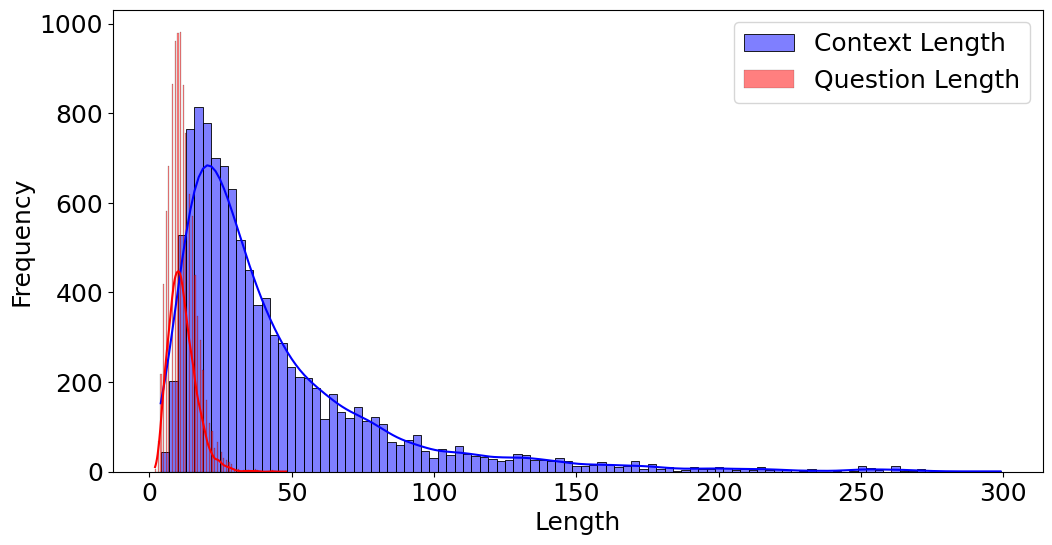}
    \caption{Length histogram of questions and contexts. Outliers longer than 300 words are excluded}
    \label{fig:enter-label}
\end{figure}

\newpage
\subsection{QA}

\begin{figure}[H]
    \centering
    \includegraphics[width=1\linewidth]{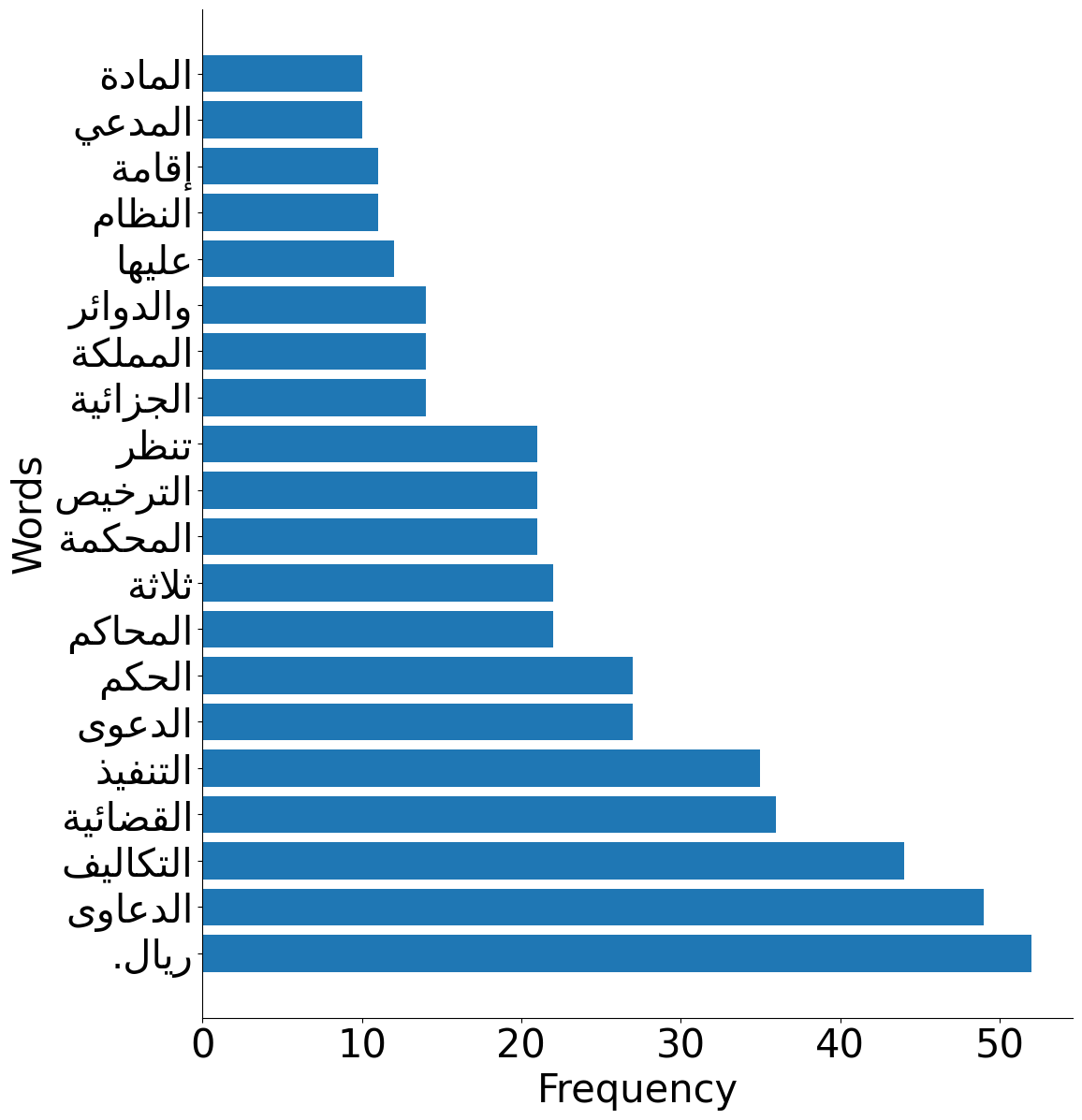}
    \caption{Top 20 frequency of words in the context of the QAs}
    \label{fig:enter-label}
\end{figure}

\begin{figure}[H]
    \centering
    \includegraphics[width=1\linewidth]{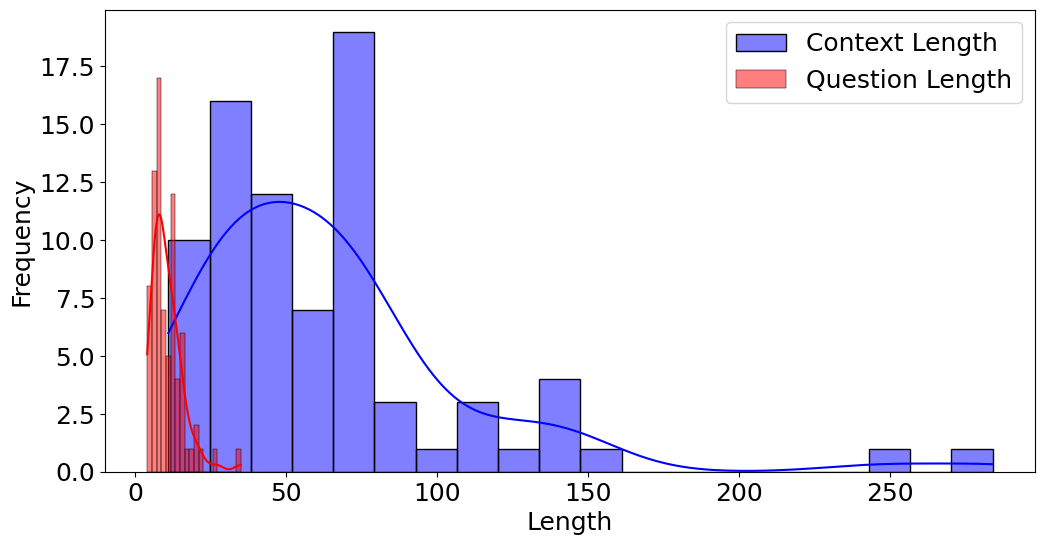}
\caption{Length histogram of questions and contexts.}
\label{fig:enter-label}
    
\end{figure}

\begin{table}[H]
    \centering
    \begin{tabular}{lccc}
        \hline
        & Min Length & Max Length & Avg Length \\
        \hline
        Context & 11 & 284 & 64.0 \\
        Question & 4 & 35 & 11.0 \\
        \hline
    \end{tabular}
    \caption{Length of context documents and questions (words)}
    \label{table:stats}
\end{table}

\onecolumn

\subsection{Arabic LegalBench}
\begin{center}
\begin{figure}[H]
    \includegraphics[width=\columnwidth]{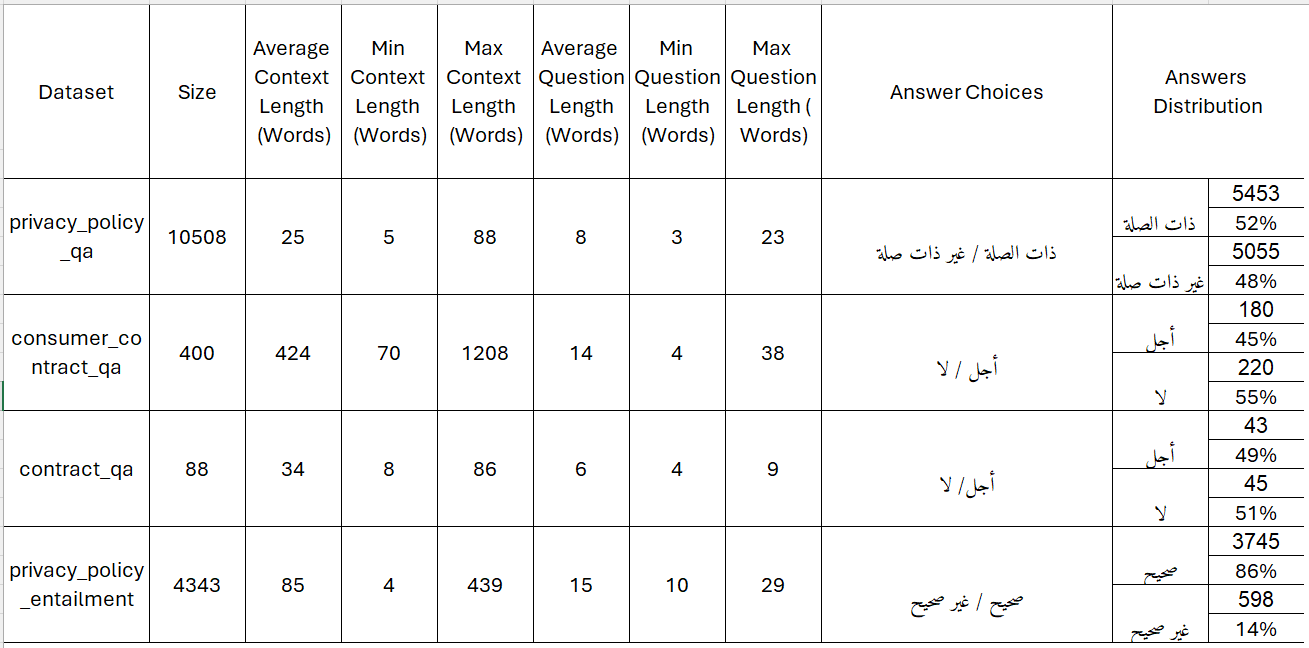}
    \caption{Summary Statistics for Arabic LegalBench}
    \label{fig:arabicLegalBenchQuant}
\end{figure}
\end{center}
\twocolumn

\subsection{Experts Performance on the Tasks}
\label{sec:experts-perfromance}

To establish a human performance baseline and provide a meaningful comparison for the LLMs' performance, a sample of the data was presented to legal experts for evaluation. These experts, with their specialized knowledge and experience, were tasked with solving the same problems that were presented to the LLMs.
Table \ref{tab:experts-results} summarize the results of the experts assessment.

\begin{table}[H]
\centering
\resizebox{\columnwidth}{!}{%
\begin{tabular}{|l|l|l|clll|}
\hline
         & \multicolumn{1}{c|}{MCQs} & NajezQA & \multicolumn{4}{c|}{Arabic LegalBench} \\ \hline
F1 Score & -                         & -       & \multicolumn{4}{c|}{0.73}              \\ \hline
Accuracy & 0.9                       & 0.65    & \multicolumn{4}{c|}{0.8}               \\ \hline
\end{tabular}%
}
\caption{Experts Answers }
\label{tab:experts-results}
\end{table}